\documentclass[10pt,journal,compsoc]{IEEEtran}

\ifCLASSOPTIONcompsoc
  \usepackage[nocompress]{cite}
\else
  \usepackage{cite}
\fi
\ifCLASSINFOpdf
  \usepackage[pdftex]{graphicx}
\else
\fi
\usepackage{amsmath,amsfonts} 
\usepackage{algorithm} 
\usepackage{algorithmic}

\usepackage{array}
\usepackage{booktabs}
\usepackage{multirow}

\usepackage[caption=false,font=footnotesize,labelfont=sf,textfont=sf]{subfig}

\usepackage{stfloats}
\usepackage{url}
\begin{document}
%
\title{Dual-Correction Adaptation Network for \\ Noisy Knowledge Transfer}
%
%
%
%

\author{Yunyun~Wang,
        Weiwen~Zheng,
        and~Songcan~Chen
\IEEEcompsocitemizethanks{\IEEEcompsocthanksitem Yunyun Wang and Weiwen Zheng are with the Jiangsu Key Laboratory of Big Data Security \& Intelligent Processing, Computer Science and Engineering, Nanjing University of Posts \& Telecommunications, Nanjing 210046, China. E-mail: wangyunyun@njupt.edu.cn, 1020041209@njupt.edu.cn.
\IEEEcompsocthanksitem Songcan Chen is with the MIIT Key Laboratory of Pattern Analysis and Machine Intelligence, Computer Science \& Technology/AI, Nanjing University of Aeronautics \& Astronautics, Nanjing 210023, China. E-mail: s.chen@nuaa.edu.cn.
}
\thanks{Corresponding Author: Songcan Chen (s.chen@nuaa.edu.cn).}
\thanks{Manuscript received April 19, 2005; revised August 26, 2015.}}

%
%

\markboth{Journal of \LaTeX\ Class Files,~Vol.~14, No.~8, August~2015}%
{Shell \MakeLowercase{\textit{et al.}}: Bare Advanced Demo of IEEEtran.cls for IEEE Computer Society Journals}
%



\IEEEtitleabstractindextext{%
\begin{abstract}
Previous unsupervised domain adaptation (UDA) methods aim to promote target learning via a \emph{single}-directional knowledge transfer from label-rich source domain to unlabeled target domain, while its reverse adaption from target to source has not jointly been considered yet so far. In fact, in some real teaching practice, a teacher helps students learn while also gets promotion from students to some extent, which inspires us to explore a \emph{dual}-directional knowledge transfer between domains, and thus propose a Dual-Correction Adaptation Network (DualCAN) in this paper. However, due to the asymmetrical label knowledge across domains, transfer from unlabeled target to labeled source poses a more difficult challenge than the common source-to-target counterpart. First, the target pseudo-labels predicted by source commonly involve noises due to model bias, hence in the reverse adaptation, they may hurt the source performance and bring a negative target-to-source transfer. Secondly, source domain usually contains innate noises, which will inevitably aggravate the target noises, leading to noise amplification across domains. To this end, we further introduce a Noise Identification and Correction (NIC) module to correct and recycle noises in both domains. To our best knowledge, this is the first naive attempt of dual-directional adaptation for noisy UDA, and naturally applicable to noise-free UDA. A theory justification is given to state the rationality of our intuition. Empirical results confirm the effectiveness of DualCAN with remarkable performance gains over state-of-the-arts, particularly for extreme noisy tasks (e.g., $\sim$+ 15$\%$ on Pw$\rightarrow$Pr and Pr$\rightarrow$Rw of Office-Home).
\end{abstract}

}

\maketitle

\IEEEdisplaynontitleabstractindextext

%
\IEEEpeerreviewmaketitle

\ifCLASSOPTIONcompsoc
\IEEEraisesectionheading{\section{Introduction}\label{sec:introduction}}
\else
\section{Introduction}
\label{sec:introduction}
\fi

%
%
%
%
\IEEEPARstart{D}{eep} neural network has achieved remarkable success in many applications, such as image classification and semantic segmentation. However, it relies on large-scared and high-quality annotated data, which is usually difficult to collect. Unsupervised domain adaptation (UDA) \cite{ben2010theory}, which aims to adopt a fully-labeled source domain to help the learning of unlabeled target domain, has attracted much attention in recent years. Most UDA methods transfer knowledge from source to target by learning domain-invariant representation across domains, mainly including discrepancy-based \cite{saito2018maximum, long2015learning} and adversarial-based methods \cite{tzeng2017adversarial, ganin2016domain, long2018conditional}. Discrepancy-based methods explicitly reduce the distribution discrepancy between domains by minimizing some distance metric, such as Maximum Mean Discrepancy (MMD) \cite{long2015learning}, Correlation Alignment (CORAL) \cite{sun2016return} and Wasserstein distance \cite{shen2018wasserstein}. Adversarial-based methods align feature distributions across domains by adversarial training between feature generator and domain discriminator \cite{ganin2016domain}, or between different classifiers \cite{saito2018maximum}.

In real UDA tasks, the source domain usually involves noises, further giving rise to noisy UDA \cite{shu2019transferable, han2020towards}. For example, source data collected from crowd-sourced platforms or internet medias will inevitably be corrupted by noises over both features and labels. The feature noise corrupts original features and may increase the difficulty of domain alignment, while label noise worsens the expected risk of classification, thus incurs misclassification of target instances. It makes previous UDA methods easy to fail in noisy environments. Recently, some researches \cite{shu2019transferable,han2020towards,yu2021divergence,zhao2020unsupervised} have been dedicated to noisy UDA learning, which can be mainly divided into two categories. One kind of methods, such as Transferable Curriculum Learning (TCL) \cite{shu2019transferable} and Robust Domain Adaptation (RDA) \cite{han2020towards}, adopts small-loss criterion to separate source instances into clean and noisy parts, then transfer source knowledge to target with only clean instances detected. The other kind, including Noisy Universal Domain Adaptation (Noisy UniDA) \cite{yu2021divergence} and Noise Resistible Mutual-Training (NRMT) \cite{zhao2020unsupervised}, uses co-learning strategy with multiple classifiers to reduce the impact of label noise in adaptation.

Those previous UDA methods all adopt a \emph{single}-directional knowledge transfer from labeled source to unlabeled target for helping the target learning. However, in some real teaching practice, a teacher helps students learn, while also gets promotion from students to some extent. Inspired by such a philosophy, the reverse adaptation from the target should intuitively be able to boost  the source learning as well, especially for weak noisy sources. To our best knowledge, however, it has not been jointly considered in UDA so far. In this paper, we attempt to explore a \emph{dual}-directional knowledge transfer between domains, and propose a Dual-Correction Adaptation Network (DualCAN) to achieve mutual promotion and cooperation across domains. However, it is worth noting that there is asymmetrical label knowledge between domains, since target domain contains much less label information than the source, thus the transfer from target to source poses a more difficult challenge than the source-to-target counterpart. First, the target pseudo-labels predicted by source commonly involve noises due to transfer and model bias, hence in the reverse adaptation, they may hurt the source performance, and consequently bring a negative target-to-source transfer. Secondly, the source domain usually contains innate noises in real tasks. It will inevitably aggravate the target noises, and incurs noise amplification across domains. To address those issues, we further introduce a crucial Noise Identification and Correction (NIC) module in DualCAN to correct the noises in both domains. After that, those corrected instances are further recycled in learning rather than simply discarded, in order for a full knowledge utilization, especially in high noisy environment.  

In implementing DualCAN, knowledge transfer iterates back and forth between source-to-target (ST) task and target-to-source (TS) task. In ST, source knowledge is adapted to generate target pseudo-labels, and the pseudo-labels are further corrected by NIC with self-supervised knowledge. In TS, the target knowledge is transferred reversely to correct source noise and further boost source learning. With such a dual-directional knowledge transfer between domains, noises in both domains are corrected collaboratively, and performances in both domains are promoted mutually. It is analogous to the philosophy that teaching benefits both the teacher and students alike. Quite naturally, dual-directional transfer can be adopted for both noisy and noise-free UDA tasks, while its learning concept is also applicable for some other learning tasks, for example, using down-stream tasks to reversely help feature learning in self-supervised learning. The main contributions of this paper are summarized as follows:

\begin{itemize}
\item{A Dual-Correction Adaptation Network (DualCAN) is proposed for noisy UDA learning. To our best knowledge, this is the first work of dual-directional adaptation to mutually promote learning and correct noises in both domains.}
\item{The noisy instances are corrected and recycled by a Noise Identification and Correction (NIC) module, in order to prevent noise amplification aross domains, and achieve a full knowledge utilization, especially in high noisy environment.}
\item{A theory justification is given to state the rationality of DualCAN. Moreover, empirical comparisons are conducted in real-world tasks under different noisy settings, in order to confirm the effectiveness of proposal.}
\end{itemize}

The rest of the paper is organized as follows, section 2 introduces the related works, section 3 gives the preliminaries, the proposed DualCAN is described in details in section 4, and the comparison results are given in section 5. Finally, section 6 is the conclusion.

\section{Related Work}
\textbf{Domain Adaptation.} In the early stage of deep domain adaptation, discrepancy-based UDA methods \cite{long2015learning,zellinger2017central} directly minimize the distribution discrepancy between domains to learn domain-invariant representations. The commonly adopted discrepancy metrics include Maximum Mean Discrepancy (MMD) \cite{long2015learning} and Correlation Alignment (CORAL) \cite{sun2016return}, Wasserstein distance \cite{shen2018wasserstein}, and Contrastive Domain Discrepancy (CDD) \cite{kang2019contrastive}. Later, inspired by the practice of adversarial learning, Ganin et al. \cite{ganin2016domain} propose a Domain Adversarial Neural Network (DANN) to reduce domain gap in feature-level. Since then, a series of researches have been proposed with adversarial learning. Liu et al. \cite{liu2016coupled} propose a generative adversarial network to learn joint distribution of multi-domain images. Long et al. \cite{long2018conditional} conduct an adversarial adaptation model using conditional distribution information. Hoffman et al. \cite{hoffman2018cycada} perform adversarial learning from both pixel-level and feature-level for domain adaptation. Another line of work \cite{french2017self,lee2013pseudo,zou2019confidence} treats domain adaptation as semi-supervised learning, and adopt a self-training framework to boost knowledge transfer. Researches in \cite{kumar2020understanding,cai2021theory} have demonstrated the effectiveness of self-training in domain adaptation under reasonable assumptions. Liu et al. \cite{liu2021cycle} improve standard self-training for cycle self-training with a two-head classifier sharing a feature extractor, so as to enforce pseudo-labels to generalize across domains. They commonly adopt a single-directional knowledge transfer from source to target, while the adaptation from target to source has not been considered, which should be able to reversely boost source learning as well, especially for weak noisy sources.

\noindent \textbf{Noisy UDA.} Ambiguous features and incorrect labels seriously influence the generalization performance of deep CNNs. Previous methods address noises mainly by designing a robust loss function \cite{miyato2018virtual,patrini2017making,reed2014training}, or filtering out noisy instances in the learning process \cite{han2018co,liu2021co,jiang2018mentornet,natarajan2013learning,sukhbaatar2014training}. When the noisy setting is introduced into UDA, the learning problem becomes much more complex, since the unreliability of target pseudo-labels will be incurred by not only domain discrepancy, but also source noises. To reduce the effect of noisy instances, one strategy is following the small-loss criterion to collect clean source data for adaptation. For example, Shu et al. \cite{shu2019transferable} propose a transferable curriculum to enhance positive transfer from clean source instances, thus mitigate negative transfer by noise. Han et al. \cite{han2020towards} improve the curriculum learning by retaining feature-corrupted data, and use a proxy distribution in adversarial network. The other uses co-learning strategy with multiple classifiers to filter out source instances with incorrect annotations. For example, Zhao et al. \cite{zhao2020unsupervised} perform mutual instance selection to select reliable instances according to peer-confidence and relationship disagreement of networks. Yu et al. \cite{yu2021divergence} study universal UDA in which target domain contains unknown classes, and optimize the divergence between two classifiers to detect noisy source instances. Chen et al. \cite{chen2021self} address noisy source-free UDA by fine-tuning the pre-trained model with both pre-generated and self-generated labels in self-supervised learning. 

Different from previous methods, we focus on both feature and label noises in UDA, as well as the recycling of noisy instances to achieve full knowledge utilization in high noisy environment. Moreover, we use a two-way knowledge transfer to mutually correct noise and promote performance in both domains, which can also be applied in noise-free UDA tasks.

\section{Preliminaries}
\subsection{Problem Statement}
In UDA, both source instances $\mathbf{X_\emph{S}} = {\{x_i\}}_i^{N_\emph{S}}$ and labels $\mathbf{Y_\emph{S}} = {\{y_i\}}_i^{N_\emph{S}}$ are given, each instance $x_i \in \mathbb{R}^d$ and $y_i \in \{1,...,K\}$, where $N_\emph{S}$ is the number of source instances, $d$ is the feature dimension, and $K$ is the class number. Target data $\mathbf{X_\emph{T}} = {\{x_i\}}_i^{N_\emph{T}}$ is an unlabeled set with $N_\emph{T}$ instances. The target distribution is commonly different from source distribution in UDA, i.e., $p_S(x) \neq p_T(x)$ or $p_S(x|y) \neq p_T(x|y)$. In real-world adaptation tasks, data collected is usually corrupted with both feature and label noises, thus we usually encounter noisy source data. Specifically, assuming a noise ratio $p_{noise}$, features of a clean instance $x_i^{CL}$ will be corrupted by noise $e$ with a probability of $p_{noise}$, i.e, $p(x_i = x_i^{CL} + e) = p_{noise}$ and $p(x_i = x_i^{CL}) = 1 - p_{noise}$. At the same time, clean label $y_i^{CL}$ is corrupted according to a noise transition matrix $ \mathcal{T} \in \mathbb{R}^{K \times K}$, where $\mathcal{T}_{kl} = p(y_i = l | y_i^{CL} = k) = p_{noise}$ denotes the probability that instances in the \emph{k-th} class is incorrectly labeled as $l$.


\begin{figure}[!t]
\centering
\subfloat[]{\includegraphics[width=2.8in]{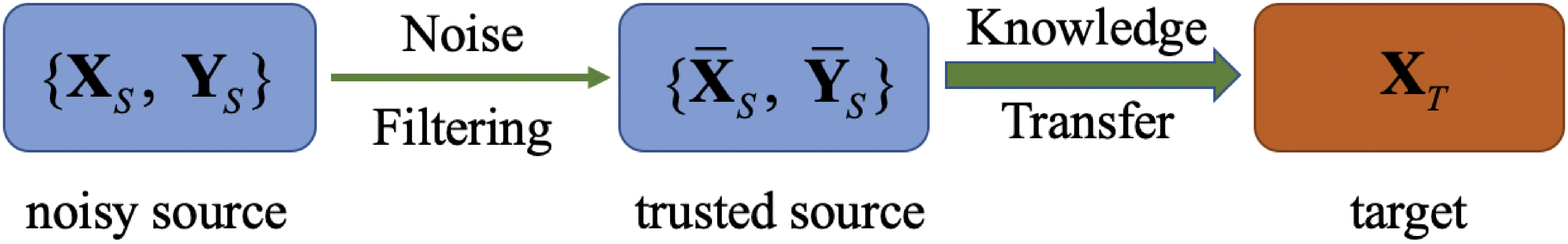}%
\label{fig1a}}
\hfil
\subfloat[]{\includegraphics[width=2.8in]{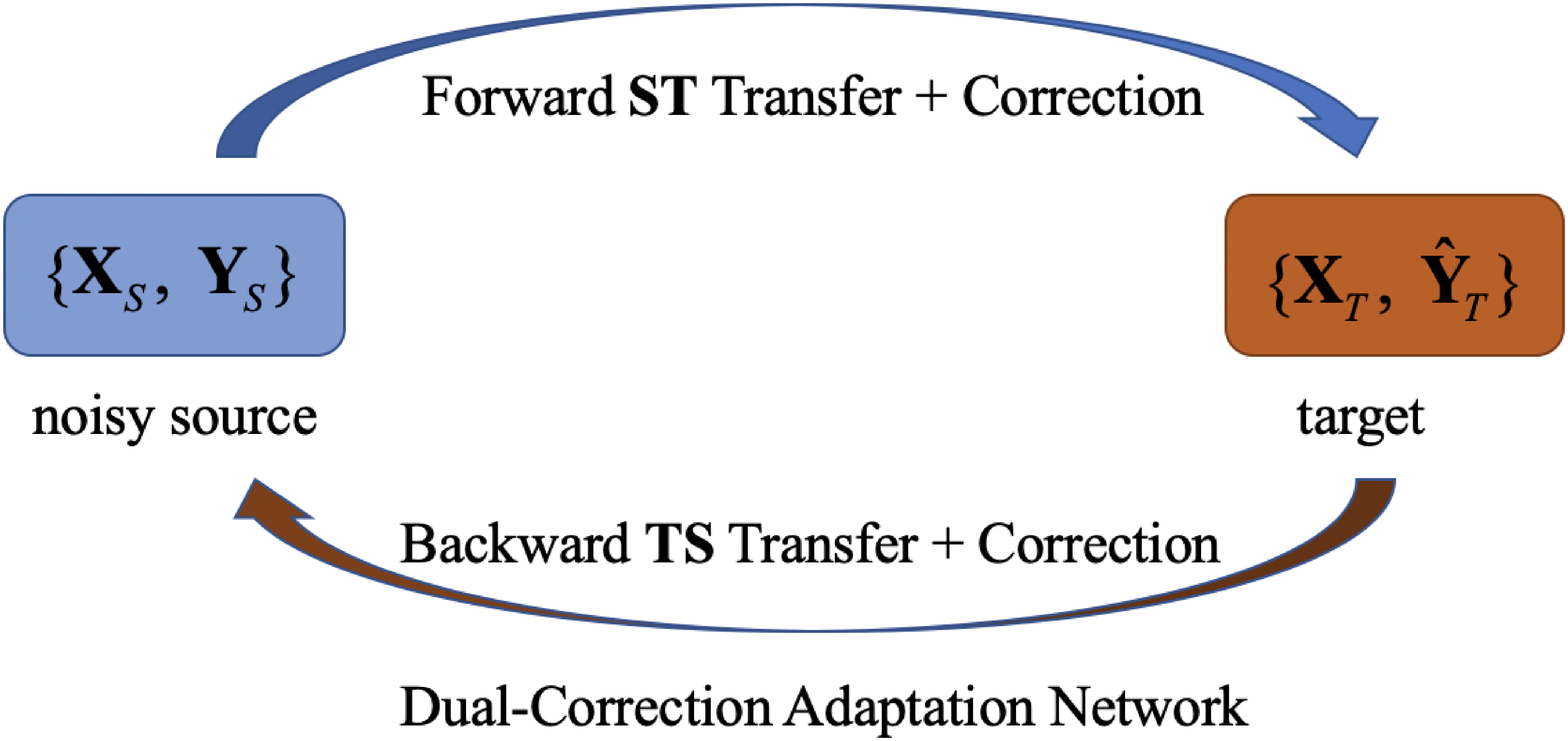}%
\label{fig1b}}
\caption{The learning process of (a) previous noisy UDA methods and (b) our propsed Dual-Correction Adaptation Network.}
\label{fig1}
\end{figure}

\subsection{Overall Concept}
Previous noisy UDA learning methods commonly adopt a signle-directional adaptation, i.e., transfer knowledge only from source domain to target to boost target learning. Moreover, they aim to filter out the noisy instances by some criterion, such as the small loss metric, or prediction discrepancy between different classifiers. Then only clean or trusted source data $\{\mathbf{\bar{X}_\emph{S}},\mathbf{\bar{Y}_\emph{S}}\}$ are selected for adaptation across domains, as shown in Fig. 1 (a). While in this paper, we propose dual-directional transfer and correction between domains, which is shown in Fig. 1 (b). Specifically, source knowledge is first transferred to target with a forward ST adaptation, together with noise correction for target pseudo-labels $\mathbf{\hat{Y}_\emph{T}}$, in order to alleviate the effect of both domain discrepancy and source noises. After that, target knowledge is reversely transferred to source by a backward TS adaptation, in order to further correct both feature and label noises in source domain for better source learning. With such dual adaptation and correction repeated, noises in both domains are corrected mutually and the learning performance of both domains can be promoted gradually.

Moreover, both feature and label noises are addressed in DualCAN, since intensive feature noise may also have negative effect to learning performance. Those noisy instances are identified and further corrected, in order for a recycling of noisy data in adaptation. The details of the proposed DualCAN will be introduced in details in the next section.

\section{The Proposed Methods}
DualCAN is designed for UDA with both feature and label noise. It learns with dual knowledge adaptation and noise correction between domains. We first introduce the network architecture of DualCAN, and then describe the dual correction and adaptation process, as well as the NIC module for noise identification and correction in separate sub-sections, respectively.

\begin{figure}[!t]
\centering
\includegraphics[width=0.95\columnwidth]{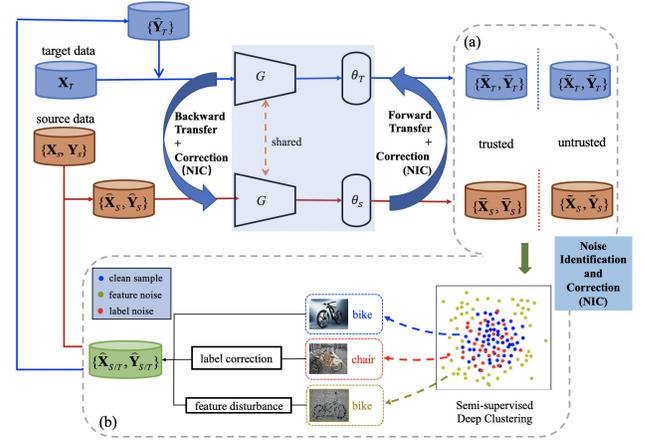}
\caption{The network architecture of DualCAN, including the dual correction and adaptation network, and the noise identification and correction module.}
\label{fig_2}
\end{figure}

\subsection{Network Architecture of DualCAN}
The network architecture of DualCAN is shown in Fig. 2. Specifically, the source and target domains share the same feature generator $G$ while different classifiers denoted as $\theta_\emph{S}$ and $\theta_\emph{T}$, respectively. First, noisy source data $\{\mathbf{X_\emph{S}},\mathbf{Y_\emph{S}}\}$ and target data $\mathbf{X_\emph{T}}$ are inputted into the network, and the initial feature generator and source classifier are obtained by source classification. Then in the ST task, source knowledge is transferred to target by assigning the target pseudo-labels, and those pseudo-labels are further corrected by the NIC module to generate $\mathbf{\overset{\frown}{Y}_\emph{T}}$. The corrected target pseudo-labels, along with self-supervised knowledge over target domain, are adopted to train target classifier. Reversely in the TS task, both source features and labels are corrected by NIC, then feature generator $G$ is updated so that target classifier can correctly classify the corrected source data $\{\mathbf{\overset{\frown}{X}_\emph{S}},\mathbf{\overset{\frown}{Y}_\emph{S}}\}$, in this way, transfer target knowledge to source and generate domain-invariant representation. Finally, a dual adaptation and correction framework is formed, so as to mutually transfer knowledge and correct noise across domains. Further in NIC, there are actually three steps. First, the small-loss criterion is adopted to roughly separate data into trusted and untrusted parts. Secondly, a semi-supervised clustering method is adopted to identify feature and label noises. Finally, those noises are further corrected for full knowledge transfer.

\subsection{Dual Correction and Adaptation}
In DualCAN, there is dual-directional knowledge transfer across domains. That is, knowledge transfer iterates between the forward source-to-target (ST) task and backward target-to-source (TS) task, so as to mutually promote learning in each domain.

The initial feature generator and source classifier are obtained by source classification. Then in the forward ST task, knowledge transfer is conducted from label-rich source domain to the label-scared target. Target pseudo-labels $\mathbf{\hat{Y}_\emph{T}} = {\{\hat{y}_i\}}_i^{N_\emph{T}}$ are first predicted by source classifier $\theta_\emph{S}$, i.e.,

\begin{equation}
\label{ex1}
\hat{y}_i = \mathop{\arg\max}\limits_{k}{_{x_i \sim \mathbf{X_\emph{T}}}} \{ f_{G,\theta_\emph{S}}(x_i)_{[k]} \}
\end{equation}

\noindent where $f_{G,\theta_\emph{S}}(x_i)_{[k]}$ is the $\emph{k-th}$ element of target prediction. Those pseudo-labels actually contain inaccurate labels or noises due to cross-domain discrepancy and source noises as well. As a result, $\mathbf{\hat{Y}_\emph{T}}$ are further corrected by the NIC module, generating $\mathbf{\overset{\frown}{Y}_\emph{T}}$ with improved label quality. Next, the target classifier $\theta_\emph{T}$ is trained with the corrected target pseudo-labels, as well as self-supervised knowledge over target domain. Specifically, for each unlabeled target instance, we conduct a consistent loss between predictions for its weakly-augmented version $x_i^1$ and a strongly-augmented version $x_i^2$, finally, the objective for learning target classifier is written as,

\begin{equation}
\label{ex2}
\begin{aligned}
\mathop{\mathrm{minimize}}\limits_{\theta_\emph{T}}{_{(x_i,\overset{\frown}{y}_i) \sim \{\mathbf{X_\emph{T}},\mathbf{\overset{\frown}{Y}_\emph{T}}\}}} & \ \mathcal{L}_{ce}(f_{G,\theta_\emph{T}}(x_i),\overset{\frown}{y}_i) \\
& + \mathcal{L}_{ce}(f_{G,\theta_\emph{T}}(x_i^1), f_{G,\theta_\emph{T}}(x_i^2)) \\
\end{aligned}
\end{equation}

\noindent where $\mathcal{L}_{ce}$ is the cross-entropy loss.

In the backward TS task, the learning goal is to transfer knowledge from corrected target domain to weak noisy source. First, both features and labels of untrusted source data are corrected by NIC, generating $\{\mathbf{\overset{\frown}{X}_\emph{S}},\mathbf{\overset{\frown}{Y}_\emph{S}}\}$ to make it closer to a noise-free distribution. Then, with the target classifier $\theta_\emph{T}$ fixed, both feature generator $G$ and source classifier $\theta_\emph{S}$ are updated by correctly classifying source instances $\{\mathbf{\overset{\frown}{X}_\emph{S}},\mathbf{\overset{\frown}{Y}_\emph{S}}\}$, in order to boost the target classifier on source domain to generate domain-invariant representation. The final learning objective is

\begin{equation}
\label{ex3}
\begin{aligned}
\mathop{\mathrm{minimize}}\limits_{G,\theta_\emph{S}}{_{(\overset{\frown}{x}_i,\overset{\frown}{y}_i) \sim \{\mathbf{\overset{\frown}{X}_\emph{S}},\mathbf{\overset{\frown}{Y}_\emph{S}}\}}} & \ \mathcal{L}_{ce}(f_{G,\theta_\emph{S}}(\overset{\frown}{x}_i),\overset{\frown}{y}_i) \\
& + \mathcal{L}_{ce}(f_{G,\theta_\emph{T}}(\overset{\frown}{x}_i),\overset{\frown}{y}_i)) \\
\end{aligned}
\end{equation}

Finally, the learning algorithm of DualCAN is shown in Algorithm 1 below, and the NIC module will be described in details in the next sub-section.

\begin{algorithm}[H]
\caption{Dual-Correction and Adaptation Network (DualCAN).}\label{alg:alg1}
\hspace*{0.02in} {\bf Input:} 
Noisy source data $\{\mathbf{X_\emph{S}},\mathbf{Y_\emph{S}}\}$ and target data $\mathbf{X_\emph{T}}$\\
\hspace*{0.02in} {\bf Output:} 
\leftline{Feature extractor $G$; Source classifier $\theta_\emph{S}$; Target}\\
classifier $\theta_\emph{T}$
\begin{algorithmic}[1]
\FOR{epoch = 0 to \emph{MaxEpoch}}
\STATE /* forward source to target task */
\STATE Generate pseudo-labels $\mathbf{\hat{Y}_\emph{T}}$ on $\mathbf{X_\emph{T}}$ with $G$ and $\theta_\emph{S}$ by Eq. 1.
\STATE Correct target pseudo-labels $\mathbf{\hat{Y}_\emph{T}}$ by the NIC module to generate $\mathbf{\overset{\frown}{Y}_\emph{T}}$.
\STATE Train target classifier $\theta_\emph{T}$ with corrected pseudo-label $\mathbf{\overset{\frown}{Y}_\emph{T}}$ and self-supervised consistency by Eq.2.
\STATE /* backward target to source task */
\STATE Correct both features and labels noises in noisy source dataset $\{\mathbf{X_\emph{S}},\mathbf{Y_\emph{S}}\}$ by NIC module to generate $\{\mathbf{\overset{\frown}{X}_\emph{S}},\mathbf{\overset{\frown}{Y}_\emph{S}}\}$.
\STATE Update feature extractor $G$ and source classifier $\theta_\emph{S}$ by Eq.3.
\ENDFOR
\end{algorithmic}
\label{alg1}
\end{algorithm}

\subsection{Noise Identification and Correction}
Due to domain discrepancy and source noise, there will be an estimation bias in the predicted target pseudo-labels, which can also be considered as noisy labels. Those target noises may hurt the source performance in the reverse adaptation. Moreover, there are actually both feature and label noises in the source domain, it will aggravate the target noises, further leading to noise amplification across domains. Thus in DualCAN, there is an extra NIC module for noise identification and correction (NIC) in each iteration. It aims to detect untrusted data, identify them as feature or label noises, and then correct them to achieve full knowledge utilization. NIC actually learns in three steps.

In the first step, both source and target data $\{\mathbf{X_\emph{S/T}},\mathbf{Y_\emph{S/T}}\}$ are roughly divided into trusted data $\{\mathbf{\overline{X}_\emph{S/T}},\mathbf{\overline{Y}_\emph{S/T}}\}$ and untrusted data $\{\mathbf{\widetilde{X}_\emph{S/T}},\mathbf{\widetilde{Y}_\emph{S/T}}\}$ by small-loss criterion, considering that clean instances have smaller losses than noisy instances, i.e.,

\begin{equation}
\label{ex4}
\begin{aligned}
x_i \in 
\begin{cases}
\mathbf{\overline{X}_\emph{S/T}},\quad &if \mathcal{L}_{ce}(f_{G,\theta_\emph{S/T}}(x_i),y_i)  \leq \gamma \\
\mathbf{\widetilde{X}_\emph{S/T}},\quad &otherwise
\end{cases}
\end{aligned}
\end{equation}

\noindent where $\gamma$ is the pre-defined threshold. If the loss of $x_i$ is smaller than $\gamma$, it will be treated as a clean instance, and a noise otherwise. In fact, only a small fraction of instances are selected into the trusted set in this step, in order to avoid introducing noise into a clean dataset. That is, the untrusted set still contains both clean instances and noises. For a source instance $x_i$, $y_i$ is its given class label and it is classified by source classifier to get the classification loss, i.e., $f_{G,\theta_\emph{S/T}}(x_i) = \theta_\emph{S}(G(x_i))$. While for target instance $x_i$, $y_i$ is the predicted pseudo-label by source classifier, and $x_i$ is classified by target classifier, i.e., $f_{G,\theta_\emph{S/T}}(x_i) = \theta_\emph{T}(G(x_i))$. Since there is no ground-truth for target instances, we actually consider that if its prediction by target classifier is inconsistent with the pseudo-label from source classifier, then the pseudo-label can be unreliable, and thus be viewed as a noisy label.

In the second step, since the trusted set contains both clean instances and noises, including feature and label noises, those different kinds of instances are further identified to reduce the negative effect. Referring to semi-supervised learning, a deep clustering method is adopted here. Specifically, the trusted data by the previous step are viewed as labeled data, while the untrusted data are treated as unlabeled data.  $K$ initial clusters $C_k \ (k=1,...,K)$ are first formed over trusted data in terms of their class labels, i.e., the \emph{k-th} cluster consists of $\overline{N}_k$ trusted instances that belong to the \emph{k-th} class,

\begin{equation}
\label{ex5}
\begin{aligned}
C_k = \{(\overline{x}_i,\overline{y}_i)|\overline{y}_i=k\}_i^{\overline{N}_k}
\end{aligned}
\end{equation}

\noindent The corresponding cluster center $\mu_k$ is computed as the average of instances in each cluster,

\begin{equation}
\label{ex6}
\begin{aligned}
\mu_k = \frac{1}{|C_k|} \sum\nolimits_{\overline{x}_i \in C_k} G(\overline{x}_i)
\end{aligned}
\end{equation}

\noindent and the cluster radius $r_k$ is the Euclidean distance between the farthest trusted instance and cluster center,

\begin{equation}
\label{ex7}
\begin{aligned}
r_k = \max\nolimits_{\overline{x}_i \in C_k} (\parallel G(\overline{x}_i) - \mu_k \parallel _2)
\end{aligned}
\end{equation}

\noindent Then for an untrusted instance $\widetilde{x}_i$, its cluster index is assigned by the nearest cluster, i.e.,

\begin{equation}
\label{ex8}
\begin{aligned}
k_i^* = \arg\min\nolimits_{k \in \{1,2,...,K\}} (\parallel G(\widetilde{x}_i) - \mu_k \parallel _2)
\end{aligned}
\end{equation}

For a source instance $\widetilde{x}_i$ in the untrusted set it will be identified as,

\begin{equation}
\label{ex9}
\begin{aligned}
\widetilde{x}_i \in 
\begin{cases}
\mathbf{X_\emph{S}^{\emph{FN}}},\quad &if \parallel G(\widetilde{x}_i) - \mu_{k_i^*} \parallel _2  > r_{k_i^*} \\
\mathbf{X_\emph{S}^{\emph{LN}}},\quad &if \parallel G(\widetilde{x}_i) - \mu_{k_i^*} \parallel _2  \leq r_{k_i^*} \ and \ \widetilde{y}_i \neq k_i^* \\
\mathbf{X_\emph{S}^{\emph{CL}}},\quad &otherwise 
\end{cases}
\end{aligned}
\end{equation}

\noindent where $\mathbf{X_\emph{S}^{\emph{FN}}}$, $\mathbf{X_\emph{S}^{\emph{LN}}}$ and $\mathbf{X_\emph{S}^{\emph{CL}}}$ denote the feature noise, label noise and clean source sets, respectively. Specifically, if $\widetilde{x}_i$ is outside the radius of the nearest cluster $C_{k_i^*}$, then $\widetilde{x}_i$ has no obvious common features with the trusted data, thus it is identified as feature noise. If $\widetilde{x}_i$ is within the radius but its label is different from the cluster center, then it has high feature similarity with trusted data but a wrong label, finally it is identified as label noise. The other instances not covered in the above two groups are assigned to clean source set with normal features and labels. For a target instance in untrusted set, its predicted pseudo-label will be identified as a label noise directly.

Finally, those noisy instances are corrected after identification. Specifically, feature noises increase the difficulty of representation learning and usually give the classifier wrong prediction results, while in many cases they do not affect the whole vision in human view. To utilize the useful information in feature noises, we take the class center $\mu_{k_i^*}$ as disturbance feature, and perform weighted disturbance on noisy features to correct instances at the feature level, i.e.,

\begin{equation}
\label{ex10}
\begin{aligned}
G(\overset{\frown}{x}_i) = (1 - \eta)G(\widetilde{x}_i) + \eta \mu_{k_i^*}
\end{aligned}
\end{equation}

\noindent where the disturbance weight $\eta \ (0 \leq \eta \leq 1)$ controls the invasion proportion of disturbance. In practice, we dynamically adjust it during training, that is, gradually reduce $\eta$ to zero as training continues. In the beginning, feature noises are corrected for better domain alignment, as training proceeds, those noises become weaker, which can be adopted to improve the model robustness. For label noises, the corrected label $\overset{\frown}{y}_i$ is directly specified by the nearest cluster label $k_i^*$, i.e., $\overset{\frown}{y}_i = k_i^*$.

The whole learning process of the NIC module is summarized in Algorithm 2 below.

\begin{algorithm}[H]
\caption{Noise Identification and Correction (NIC).}\label{alg:alg2}
\hspace*{0.02in} {\bf Input:} 
Noisy dataset $\{\mathbf{X_\emph{S}},\mathbf{Y_\emph{S}}\}$ and $\{\mathbf{X_\emph{T}},\mathbf{\hat{Y}_\emph{T}}\}$ ; Feature extractor $G$ ; Disturbance weight $\lambda$\\
\hspace*{0.02in} {\bf Output:} 
\leftline{Corrected dataset $\{\mathbf{\overset{\frown}{X}_\emph{S}},\mathbf{\overset{\frown}{Y}_\emph{S}}\}$ and $\{\mathbf{\overset{\frown}{X}_\emph{T}},\mathbf{\overset{\frown}{Y}_\emph{T}}\}$}
\begin{algorithmic}[1]
\STATE Split noisy data into trusted $\{\mathbf{\overline{X}_\emph{S/T}},\mathbf{\overline{Y}_\emph{S/T}}\}$ and untrusted $\{\mathbf{\widetilde{X}_\emph{S/T}},\mathbf{\widetilde{Y}_\emph{S/T}}\}$ parts according to Eq. 4.
\STATE Initialize clusters $C_k \ (k=1,...,K)$ with class center $\mu_k$ and radius $r_k$ by Eq. 5-7.
\FOR{all $(\widetilde{x}_i, \widetilde{y}_i) \in \{\mathbf{\widetilde{X}_\emph{S}},\mathbf{\widetilde{Y}_\emph{S}}\}$}
\STATE Find its nearest cluster class $k_i^*$ by Eq. 8.
\IF{$\parallel G(\widetilde{x}_i) - \mu_{k_i^*} \parallel _2  > r_{k_i^*}$}
\STATE Identified as feature noise and perform weighted disturbance by Eq. 10.
\ELSIF{$\parallel G(\widetilde{x}_i) - \mu_{k_i^*} \parallel _2  \leq r_{k_i^*} \ and \ \widetilde{y}_i \neq k_i^*$}
\STATE Identified as label noise and correct the label by $\overset{\frown}{y}_i = k_i^*$.
\ENDIF
\STATE Assign instances to the corresponding cluster: $C_{k_i^*} = C_{k_i^*} \bigcup \{(\overset{\frown}{x}_i, \overset{\frown}{y}_i)\}$
\ENDFOR
\FOR{all $(\widetilde{x}_i, \widetilde{y}_i) \in \{\mathbf{\widetilde{X}_\emph{T}},\mathbf{\widetilde{Y}_\emph{T}}\}$}
\STATE Correct the label by $\overset{\frown}{y}_i = k_i^*$.
\ENDFOR
\end{algorithmic}
\label{alg2}
\end{algorithm}

\subsection{Theory Justification}
In this sub-section, we give a theory justification for DualCAN, based on a noisy version of generalization bound for target domain.

\noindent \textbf{Proposition 1} \cite{han2020towards} For any hypothesis $h \in \mathcal{H}$, the bound of target expected risk $\epsilon_T(h)$ in noisy environments is given by

\begin{equation}
\label{ex11}
\begin{aligned}
\epsilon_T(h) \leq \epsilon_{S}(h) + | \epsilon_{S}(h,h^*) - \epsilon_T(h,h^*)| + \lambda
\end{aligned}
\end{equation}

\noindent where $\lambda = \epsilon_{S}(h^*,f_S) + \epsilon_T(h^*,f_T)$ is the ideal combined error of $h^*$, $f_S$ and $f_T$ are the true labeling functions for source and target domains, respectively, and,

\begin{equation}
\label{ex12}
\begin{aligned}
h^* = \mathop{\arg\min}\limits_{h \in \mathcal{H}}{} \epsilon_{S}(h,f_S) + \epsilon_T(h,f_T)
\end{aligned}
\end{equation}

From proposition 1, The target risk in noisy UDA is bounded by three items. The first item is the empirical risk of noisy source data. In DualCAN, a reverse knowledge transfer from target to source is adopted, the aligned target data can actually be treated as augmented source data after adaptation, thus the target data structure can help correct source noises and reduce source risk. At the same time, noisy source instances are corrected and recycled in DualCAN, which can further reduce the source risk, especially under extremely noisy environment. In fact, a better source learning guarantees better target learning, and reversely, it helps correct source noises and promote source performance as well, finally a virtuous circle is formed. The second item is the distribution discrepancy across noisy source domain and target domain, and the feature noise in source domain will affect the discrepancy \cite{han2020towards}. Considering that, feature noises in source domain are also addressed in DualCAN for better domain alignment. 

Further, for any labeling functions $f_1$, $f_2$ and $f_3$, we have \cite{zhu2020deep}

\begin{equation}
\label{ex14}
\begin{aligned}
\epsilon(f_1,f_2) \leq \epsilon(f_1,f_3) + \epsilon(f_2,f_3)
\end{aligned}
\end{equation}

\noindent Then we have

\begin{equation}
\label{ex15}
\begin{aligned}
\lambda = & \mathop{\min}\limits_{h \in \mathcal{H}}{} \epsilon_{S}(h,f_S) + \epsilon_T(h,f_T) \\
\leq & \mathop{\min}\limits_{h \in \mathcal{H}}{} \epsilon_{S}(h,f_S) + \epsilon_T(h,f_S) + \epsilon_T(f_S,f_T) \\
\end{aligned}
\end{equation}

The first and second items denote the disagreement between $h$ and the source labeling function $f_S$ on soure and target data, respectively. The third item is the gap of source and target functions over target instances, which is actually fixed with given data. Since the source instances are labeled, the first item can be very small. As a result, we focus on the second item $\epsilon_{T}(h,f_S) = \mathbb{E}_{x \sim \mathbf{X_\emph{T}}}[l(h(x),f_S(x))]$ for smaller $\lambda$, where $l(\cdot,\cdot)$ is the 0-1 loss. By reversly transfer knowledge from target to source, it is helpful to reduce the discrepancy between $h$ and $f_S$ over the target data.

\begin{table*}
\label{tab1}
  \caption{Classification Accuracy ($\%$) on Office-31 with 40$\%$ Corruption of Label, Feature and Mixed.}
  \centering
  \resizebox{\textwidth}{17.8mm}{
    \begin{tabular}{c|ccccccc|ccccccc|ccccccc}
    \specialrule{0.1em}{0pt}{3pt}
    \multirow{2}*{Method} & \multicolumn{7}{c|}{Label Corruption} & \multicolumn{7}{c|}{Feature Corruption} & \multicolumn{7}{c}{Mixed Corruption} \\
    \cmidrule{2-22}
    ~ & A$\rightarrow$W & W$\rightarrow$A & A$\rightarrow$D & D$\rightarrow$A & W$\rightarrow$D & D$\rightarrow$W & Avg & A$\rightarrow$W & W$\rightarrow$A & A$\rightarrow$D & D$\rightarrow$A & W$\rightarrow$D & D$\rightarrow$W & Avg & A$\rightarrow$W & W$\rightarrow$A & A$\rightarrow$D & D$\rightarrow$A & W$\rightarrow$D & D$\rightarrow$W & Avg \\
    \specialrule{0.05em}{3pt}{3pt}
    ResNet & 47.2 & 33.0 & 47.1 & 31.0 & 68.0 & 58.8 & 47.5 & 70.2 & 55.1 & 73.0 & 55.0 & 94.5 & 87.2 & 72.5 & 58.8 & 39.1 & 69.3 & 37.7 & 75.2 & 75.5 & 59.3 \\
    SPL & 72.6 & 50.0 & 75.3 & 38.9 & 83.3 & 64.6 & 64.1 & 75.8 & 59.7 & 75.7 & 56.7 & 93.9 & 87.8 & 74.9 & 77.3 & 57.5 & 78.4 & 47.5 & 93.4 & 83.5 & 72.9 \\
    MentorNet & 74.4 & 54.2 & 75.0 & 43.2 & 85.9 & 70.6 & 67.2 & 76.0 & 60.3 & 75.5 & 59.1 & 93.4 & 89.9 & 75.7 & 76.8 & 59.5 & 78.2 & 52.3 & 94.4 & 89.0 & 75.0 \\
    RTN & 64.6 & 56.2 & 76.1 & 49.0 & 82.7 & 71.7 & 66.7 & 81.0 & 64.6 & 81.3 & 62.3 & 95.2 & 91.0 & 79.2 & 76.7 & 56.9 & 84.1 & 56.4 & 93.0 & 86.7 & 75.6 \\
    DANN & 61.2 & 46.2 & 57.4 & 42.4 & 74.5 & 62.0 & 57.3 & 71.3 & 54.1 & 69.0 & 54.1 & 84.5 & 84.6 & 69.6 & 69.7 & 50.0 & 69.5 & 49.1 & 80.1 & 79.7 & 66.4 \\
    MDD & 74.7 & 55.1 & 76.7 & 54.3 & 89.2 & 81.6 & 71.9 & 92.9 & 66.8 & 88.0 & 70.9 & 99.8 & 96.6 & 85.8 & 88.7 & 63.1 & 81.9 & 68.5 & 94.6 & 89.3 & 81.0 \\
    CST & 77.1 & 57.0 & 77.9 & 58.9 & 90.3 & 82.5 & 73.9 & 93.7 & 67.1 & 88.9 & 71.2 & 99.8 & 97.0 & 86.3 & 90.8 & 64.4 & 82.9 & 68.3 & 94.8 & 90.2 & 81.9 \\
    TCL & 82.0 & 65.7 & 83.3 & 60.5 & 90.8 & 77.2 & 76.6 & 84.9 & 62.3 & 83.7 & 64.0 & 93.4 & 91.3 & 79.9 & 87.4 & 64.6 & 83.1 & 62.2 & 99.0 & 92.7 & 81.5 \\
    RDA & 89.7 & 67.2 & 92.0 & 65.5 & 96.0 & 92.7 & 83.6 & \textbf{95.1} & 68.4 & 89.4 & 72.4 & 99.8 & 97.8 & 87.2 & 93.1 & 69.5 & 92.0 & 71.5 & 99.0 & 93.1 & 86.4 \\
    \specialrule{0.05em}{3pt}{3pt}
    DualCAN & \textbf{94.2} & \textbf{71.6} & \textbf{93.4} & \textbf{75.3} & \textbf{98.3} & \textbf{95.7} & \textbf{88.1} & 94.8 & \textbf{71.9} & \textbf{93.2} & \textbf{76.7} & \textbf{100.0} & \textbf{98.6} & \textbf{89.4} & \textbf{95.0} & \textbf{70.9} & \textbf{93.4} & \textbf{76.5} & \textbf{99.8} & \textbf{97.4} & \textbf{88.8} \\
    \specialrule{0.1em}{3pt}{0pt}
    \end{tabular}
  }
\end{table*}

\begin{table*}
\caption{Classification Accuracy ($\%$) on Office-Home with 160$\%$ Mixed Corruption and Bing-Caltech with Native Noises.}
\label{tab2}
\centering
\resizebox{\textwidth}{21.5mm}{
  \begin{tabular}{c|cccccccccccc|c|c}
  \specialrule{0.1em}{0pt}{3pt}
  \multirow{2}*{Method} & \multicolumn{13}{c|}{Office-Home} & Bing-Caltech \\
  \cmidrule{2-15}
  ~ & Ar$\rightarrow$Cl & Ar$\rightarrow$Pr & Ar$\rightarrow$Rw & Cl$\rightarrow$Ar & Cl$\rightarrow$Pr & Cl$\rightarrow$Rw & Pr$\rightarrow$Ar & Pr$\rightarrow$Cl & Pr$\rightarrow$Rw & Rw$\rightarrow$Ar & Rw$\rightarrow$Cl & Rw$\rightarrow$Pr & Avg & B$\rightarrow$C \\
  \specialrule{0.05em}{3pt}{3pt}
  ResNet & 8.0 & 11.2 & 13.2 & 7.4 & 12.1 & 12.1 & 8.9 & 8.9 & 14.9 & 10.7 & 9.4 & 17.3 & 11.2 & 74.4 \\
  SPL & 11.5 & 15.2 & 16.9 & 8.8 & 15.9 & 15.3 & 17.6 & 15.2 & 22.6 & 15.5 & 11.9 & 17.2 & 15.3 & 75.3 \\
  MentorNet & 13.8 & 15.5 & 17.3 & 9.4 & 16.0 & 15.9 & 17.4 & 16.3 & 24.2 & 16.0 & 11.1 & 21.3 & 16.2 & 75.6 \\
  RTN & 8.5 & 10.9 & 14.7 & 6.9 & 11.1 & 12.8 & 10.5 & 9.8 & 15.2 & 12.0 & 9.8 & 17.5 & 11.6 & 75.8 \\
  DANN & 7.3 & 12.6 & 13.0 & 7.8 & 12.5 & 9.8 & 8.0 & 9.1 & 15.5 & 11.7 & 9.1 & 18.8 & 11.3 & 72.3 \\
  MDD & 8.8 & 11.9 & 19.7 & 13.5 & 13.9 & 10.9 & 11.4 & 10.9 & 21.5 & 12.8 & 11.5 & 17.9 & 13.7 & 78.9 \\
  CST & 8.6 & 12.0 & 19.3 & 11.2 & 13.4 & 13.0 & 11.8 & 11.4 & 19.7 & 13.3 & 12.0 & 17.7 & 13.6 & 79.5 \\
  TCL & 12.0 & 19.1 & 22.4 & 14.0 & 25.6 & 21.2 & 18.7 & 20.9 & 33.9 & 26.2 & 22.7 & 27.4 & 22.0 & 79.0 \\
  RDA & 17.8 & 25.2 & 29.4 & 17.7 & 28.8 & 27.0 & 21.5 & 21.3 & 34.1 & 25.5 & 22.1 & 34.3 & 25.4 & 81.7 \\
  \specialrule{0.05em}{3pt}{3pt}
  DualCAN & \textbf{31.1} & \textbf{39.4} & \textbf{44.1} & \textbf{30.6} & \textbf{42.0} & \textbf{40.8} & \textbf{35.9} & \textbf{36.2} & \textbf{50.3} & \textbf{39.9} & \textbf{37.9} & \textbf{59.0} & \textbf{40.6} & \textbf{82.4} \\
  \specialrule{0.1em}{3pt}{0pt}
  \end{tabular}
}
\end{table*}

In summary, by dual-directional knowledge transfer between domains, as well as correction and recycling for both feature and label noises, it can be expected to achieve better performance by DualCAN.

\section{Experiments}
In this section, we evaluate the proposed DualCAN on three vision datasets, against the state-of-art UDA and noisy UDA methods.

\subsection{Setup}
\textbf{Datasets.} \emph{Office-31} \cite{saenko2010adapting} is a standard domain adaptation dataset containing 4652 images with 31 classes. It consists of three domains: \emph{Amazon} (\textbf{A}), \emph{Webcam} (\textbf{W}) and \emph{DSLR} (\textbf{D}). \emph{Office-Home} \cite{venkateswara2017deep} has 15599 images with 65 classes. It contains 4 domains with large domain gaps: \emph{Artistic} (\textbf{Ar}), \emph{Clip Art} (\textbf{Cl}), \emph{Product} (\textbf{Pr}) and \emph{Real-World} (\textbf{Rw}). To introduce noise in those two clean datasets, we follow the protocol in \cite{han2020towards} to create corrupted counterparts in 3 different ways: label corruption, feature corruption, and mixed corruption. Label corruption changes the label of each image to other random classes uniformly and equally with probability $p_{noise}$. Feature corruption corrupts image pixels with probability of $p_{noise}$ by Gaussian blur and salt-and-pepper noise. Mixed corruption refers that each image is corrupted by label corruption and feature corruption with probability $p_{noise}/2$ independently. \emph{Bing-Caltech} \cite{bergamo2010exploiting} is a real noisy dataset consisting of \emph{Bing} (\textbf{B}) and \emph{Caltech-256} (\textbf{C}). The Bing dataset is created by collecting images from the Bing image search engine with class labels in Caltech-256, which natually contains label and feature noises. We take Bing as the noisy source domain and Caltech-256 as the clean target domain.

\noindent \textbf{Benchmark methods.} We compare DualCAN with state-of-the-art methods: ResNet-50 \cite{he2016deep}, Self-Paced Learning (SPL) \cite{kumar2010self}, MentorNet \cite{jiang2018mentornet}, Deep Adaptation Network (DAN) \cite{long2015learning}, Residual Transfer Network (RTN) \cite{long2016unsupervised}, Domain Adversarial Neural Network (DANN) \cite{ganin2016domain}, Conditional adversarial Domain Adaptation Network (CDAN) \cite{long2018conditional}, Margin Disparity Discrepancy (MDD) \cite{zhang2019bridging}, Cycle Self-Training (CST) \cite{liu2021cycle}, Transferable Curriculum Learning (TCL) \cite{shu2019transferable}, and Robust Domain Adaptation (RDA) \cite{han2020towards}. SPL and MentorNet are label noise processing methods, TCL and RDA are noisy domain adaptation methods, while the others are standard UDA methods.

\noindent \textbf{Implementation.} We use all labeled source and unlabeled target instances for training following the standard protocols in UDA \cite{ganin2016domain}, implement both DualCAN and comparison methods in Pytorch. We use ResNet-50 pretrained on ImageNet \cite{russakovsky2015imagenet} as feature extractors, and a fully connected bottleneck layer before the classification layer. We set the sample selection pre-training epoch to 30. The small-loss threshold $\gamma$ is usually determined by empirical or prior knowledge of noise in previous methods \cite{shu2019transferable,han2020towards}. We fix the separation ratio $p=0.08$ to set $\gamma$ as the loss of the $(N \times p)$\emph{-th} instance in most tasks. Following standard protocol in \cite{he2016deep}, we set the initial learning rate to $2e-3$, and decay the learning rate by 0.1 in each 30 of the 90 epochs.

\begin{table*}
\caption{Classification Accuracy ($\%$) on Office-Home for noise-free unsupervised domain adaptation.}
\label{tab3}
\centering
\resizebox{\textwidth}{16.5mm}{
  \begin{tabular}{c|cccccccccccc|c}
  \specialrule{0.1em}{0pt}{3pt}
  Method & Ar$\rightarrow$Cl & Ar$\rightarrow$Pr & Ar$\rightarrow$Rw & Cl$\rightarrow$Ar & Cl$\rightarrow$Pr & Cl$\rightarrow$Rw & Pr$\rightarrow$Ar & Pr$\rightarrow$Cl & Pr$\rightarrow$Rw & Rw$\rightarrow$Ar & Rw$\rightarrow$Cl & Rw$\rightarrow$Pr & Avg \\
  \specialrule{0.05em}{3pt}{3pt}
  ResNet & 34.9 & 50 & 58 & 37.4 & 41.9 & 46.2 & 38.5 & 31.2 & 60.4 & 53.9 & 51.2 & 59.9 & 46.1 \\
  DAN & 43.6 & 57 & 67.9 & 45.8 & 56.5 & 60.4 & 44 & 43.6 & 67.7 & 63.1 & 51.5 & 74.3 & 56.3 \\
  DANN & 45.6 & 59.3 & 70.1 & 47 & 58.5 & 60.9 & 46.1 & 43.7 & 68.5 & 63.2 & 51.8 & 76.8 & 57.6 \\
  CDAN & 50.7 & 70.6 & 76 & 57.6 & 70 & 70 & 57.4 & 50.9 & 77.3 & 70.9 & 56.7 & 81.6 & 65.8 \\
  MDD & 54.9 & 73.7 & 77.8 & 60 & 71.4 & 71.8 & 61.2 & 53.6 & 78.1 & 72.5 & 60.2 & 82.3 & 68.1 \\
  CST & 59 & 79.6 & \textbf{83.4} & 68.4 & 77.1 & 76.7 & 68.9 & 56.4 & \textbf{83} & 75.3 & 62.2 & \textbf{85.1} & 73 \\
  \specialrule{0.05em}{3pt}{3pt}
  DualCAN & \textbf{61.7} & \textbf{79.8} & 83.1 & \textbf{69.4} & \textbf{79.8} & \textbf{77.1} & \textbf{70.5} & \textbf{57.2} & 81.7 & \textbf{77.9} & \textbf{62.5} & 83.9 & \textbf{73.7} \\
  \specialrule{0.1em}{3pt}{0pt}
  \end{tabular}
}
\end{table*}

\begin{table}
\label{tab4}
  \caption{Classification Accuracy ($\%$) on Office-31 for noise-free unsupervised domain adaptation.}
  \centering
  \resizebox{\linewidth}{16mm}{
    \begin{tabular}{c|cccccc|c}
    \specialrule{0.1em}{0pt}{3pt}
    Method & A$\rightarrow$W & W$\rightarrow$A & A$\rightarrow$D & D$\rightarrow$A & W$\rightarrow$D & D$\rightarrow$W & Avg \\
    \specialrule{0.05em}{3pt}{3pt}
    ResNet & 68.4 & 60.7 & 68.9 & 62.5 & 99.3 & 96.7 & 76.1 \\
    DAN & 80.5 & 62.8 & 78.6 & 63.6 & 99.6 & 97.1 & 80.4 \\
    DANN & 82 & 67.4 & 79.7 & 68.2 & 99.1 & 96.9 & 82.2 \\
    CDAN & 94.1 & 69.3 & 92.9 & 71 & \textbf{100} & 98.6 & 87.7 \\
    MDD & 94.5 & 72.2 & 93.5 & 74.6 & \textbf{100} & 98.4 & 88.9 \\
    CST & 89.3 & 72.7 & 93.2 & 77 & \textbf{100} & \textbf{99.2} & 88.6 \\
    \specialrule{0.05em}{3pt}{3pt}
    DualCAN & \textbf{95.1} & \textbf{73.1} & \textbf{93.6} & \textbf{78.2} & \textbf{100} & 99 & \textbf{89.8} \\
    \specialrule{0.1em}{3pt}{0pt}
    \end{tabular}
  }
\end{table}

\subsection{Results}
\textbf{Noisy UDA.} Tables 1 show the results on Office-31 under 40$\%$ label corruption, feature corruption, and mixed corruption, respectively. In mixed corruption case, 40$\%$ means 20$\%$ feature corruption and 20$\%$ label corruption independently. From those tables, we can find that

\begin{itemize}
\item{Through addressing label noise in learning, both SPL and MentorNet obtain better performance than ResNet, however, since they directly apply source classifier over target instances without taking the domain discrepancy into account, the target performance can be further boosted.}
\item{The UDA methods, such as MDD and CST, consider the domain alignment in learning, whereas there is noise in source domain, and the target learning will be affected by the source noise, leading to negative transfer. As a result, their performance can be further boosted, and it is important to correct the source noise in domain adaptation for better knowledge transfer.}
\item{TCL and RDA consider both source noise and domain discrepancy in learning. They filter out noise instances with small-loss criterion to reduce the negative impact of noise in adaptation, thus achieve better performance than those noise processing and UDA methods.}
\item{Our DualCAN outperforms the other methods in almost all cases, including TCL and RDA, indicating the effectiveness of the proposed dual correction and adaptation.}
\end{itemize}

\begin{table}
\caption{Ablation study on Office-31 with 40$\%$ Mixed Corruption.}
\label{tab5}
\centering
\resizebox{\linewidth}{13.5mm}{
  \begin{tabular}{c|cccccc|c}
  \specialrule{0.1em}{0pt}{3pt}
  \multirow{2}*{\normalsize Method} & \multicolumn{7}{c}{Office-31 40$\%$ Mixed Corruption} \\
  \cmidrule{2-8}
  ~ & A$\rightarrow$W & W$\rightarrow$A & A$\rightarrow$D & D$\rightarrow$A & W$\rightarrow$D & D$\rightarrow$W & Avg \\
  \specialrule{0.05em}{3pt}{3pt}
  w/o feature correction & 94.0 & 70.7 & 92.9 & 76.0 & 99.2 & 96.6 & 88.2 \\
  w/o label correction & 86.1 & 63.6 & 84.2 & 68.5 & 91.6 & 88.7 & 80.5 \\
  w/o source correction & 85.0 & 62.5 & 83.4 & 67.6 & 90.5 & 87.5 & 79.4 \\
  w/o target correction & 90.2 & 67.1 & 89.2 & 73.8 & 96.2 & 93.6 & 85.0 \\
  \specialrule{0.05em}{3pt}{3pt}
  DualCAN & \textbf{95.0} & \textbf{70.9} & \textbf{93.4} & \textbf{76.5} & \textbf{99.8} & \textbf{97.4} & \textbf{88.8} \\
  \specialrule{0.1em}{3pt}{0pt}
  \end{tabular}
}
\end{table}

Table 2 shows the results in extremely noisy environment on Office-Home with 160$\%$ mixed corruption, including 80$\%$ label corruption and 80$\%$ feature corruption independently, as well as naive noisy environment over Bing-Caltech. From this table, we have a similar conclusion that DualCAN outperforms the other methods, since standard UDA methods usually overfit the source noise, while standard noise processing methods do not take the discrepancy across domains into account. Further, DualCAN adopts a dual correction and adaptation strategy, instead of only allowing clean samples to participate training, the noise instances are corrected and recycled in DualCAN. Finally, DualCAN achieves better performance than both TCL and RDA, demonstrating the robustness of our proposal in high noisy environment.

\noindent \textbf{Noise-free UDA.} Besides noisy UDA, dual-directional transfer is also applicable to noise-free UDA, thus we give the performance comparison in noise-free UDA tasks in Table 3 and 4. From those tables, it can be found that DualCAN achieves competitive performance compared with the state-of-art UDA methods. Specifically, DualCAN obtains the best performance over 5 out of the 9 adaptation tasks over Office-31, and over 9 out of the 12 tasks over Office-Home. Moreover, it achieves the best average performance on both datasets. As a result, a dual-directional knowledge transfer across domains can achieve competitive performance in noise-free UDA as well.

\begin{figure*}[!t]
\centering
\subfloat[]{\includegraphics[scale=0.3]{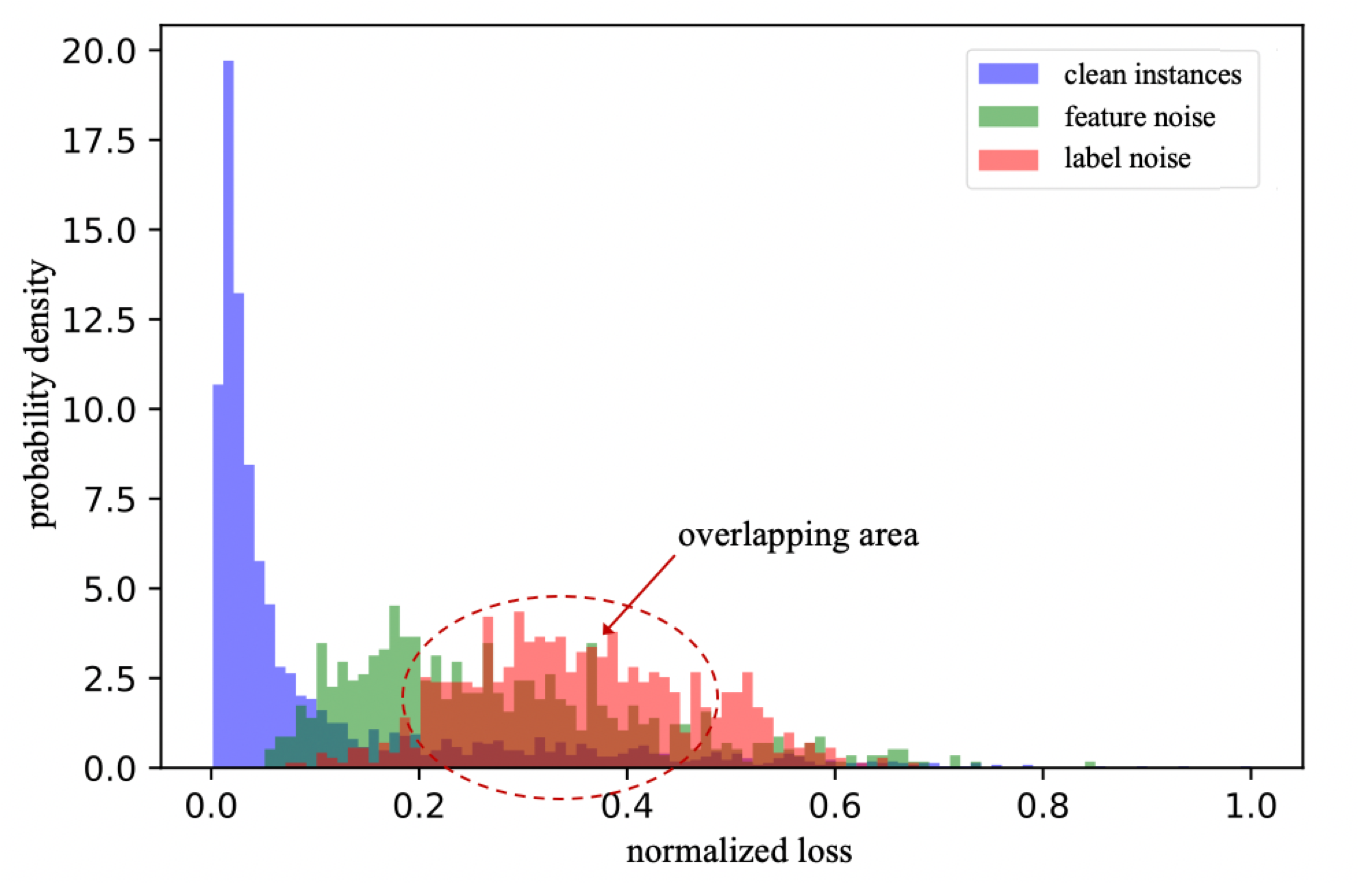}
\label{fig3a}}
\subfloat[]{\includegraphics[scale=0.3]{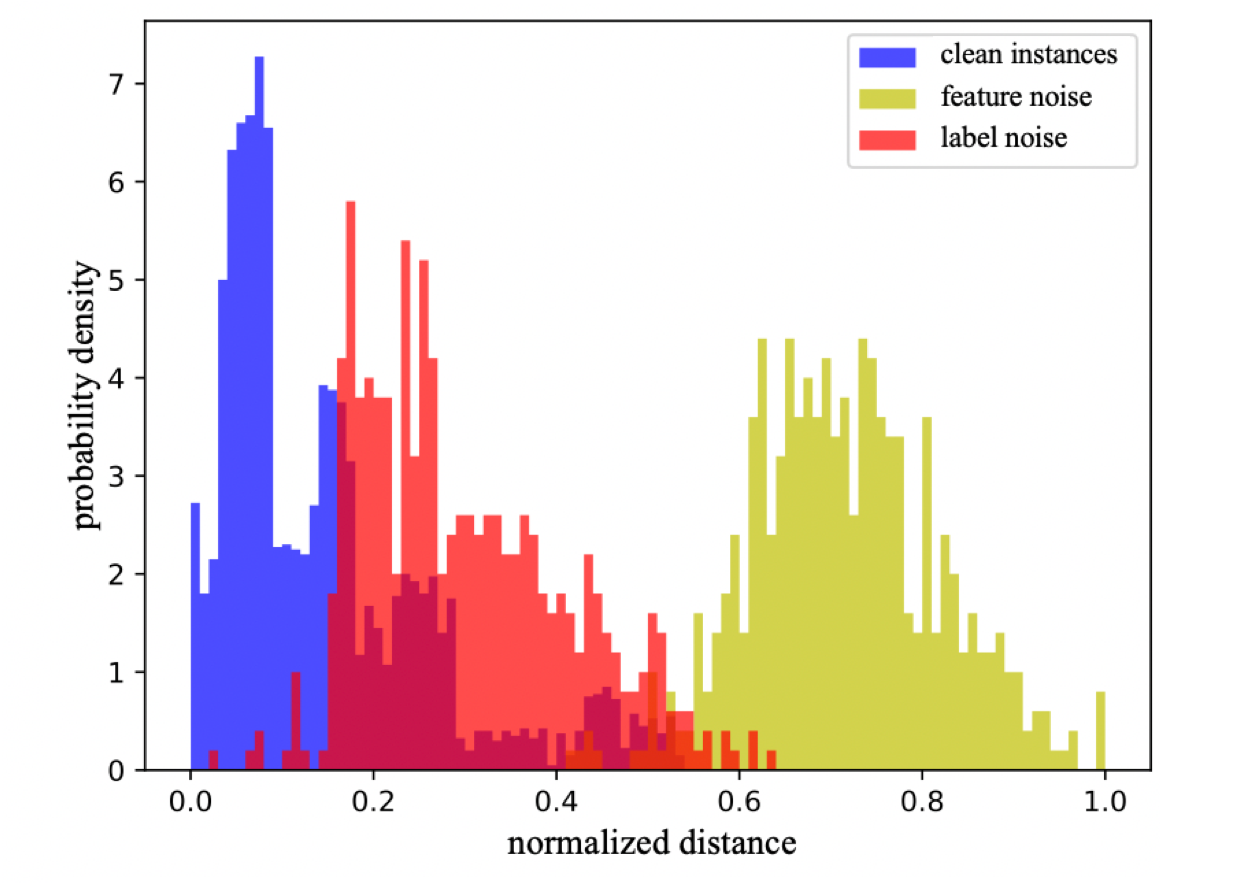}
\label{fig3b}}
\subfloat[]{\includegraphics[scale=0.3]{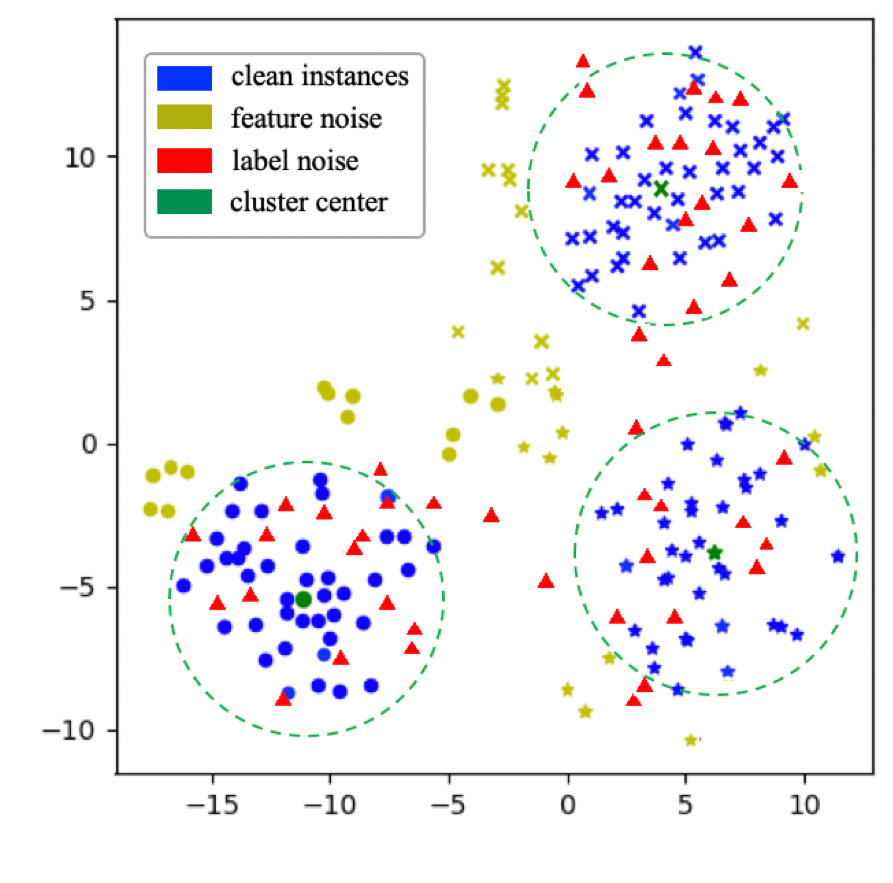}
\label{fig3c}}
\caption{(a): The loss distributions of instances. (b): The distance distribution of instances by clustering. (c): Visualization of deep clusters for random three classes in Office-Home Product. The green dashed circles represent the noise determination range with green cluster centers and $r_k$ as radius.}
\label{fig3}
\end{figure*}

\begin{figure}[t]
\centering
\includegraphics[width=0.9\columnwidth]{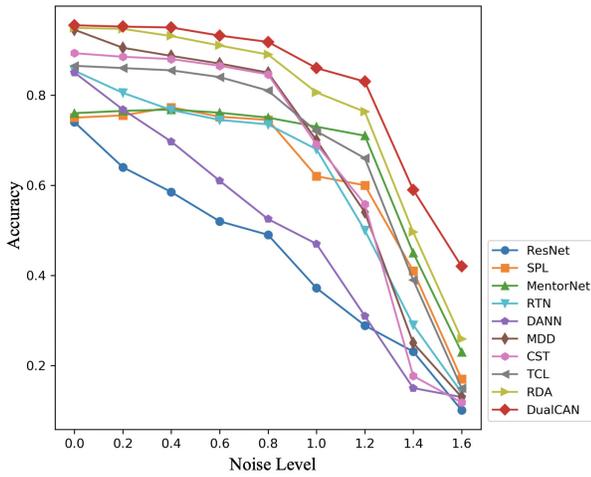}
\caption{Analysis of various methods in different mixed noise levels.}
\label{fig4}
\end{figure}

\subsection{Analysis}
\textbf{Ablation Study.} In order to investigate the contribution of each component in DualCAN better, we perform an ablation study on Office-31 with 40$\%$ mixed corruption by deleting some components in DualCAN, and the results are reported in Table 5. From Table 5, it can be found that abandoning either label correction or feature correction leads to performance degradation, thus it is reasonable to address both label and feature noises in learning. At the same time, label noise has a greater impact to performance, demonstrating the necessary for different treatment of label and feature noises. Furthermore, the performance of DualCAN is better than source correction or target correction alone, which indicates that the dual correction is indeed helpful for alleviating the negative effects of noisy instances, and improving the quality of pseudo-labels, thus boosts both source and target learning.

\begin{figure}[!t]
\centering
\includegraphics[width=0.95\columnwidth]{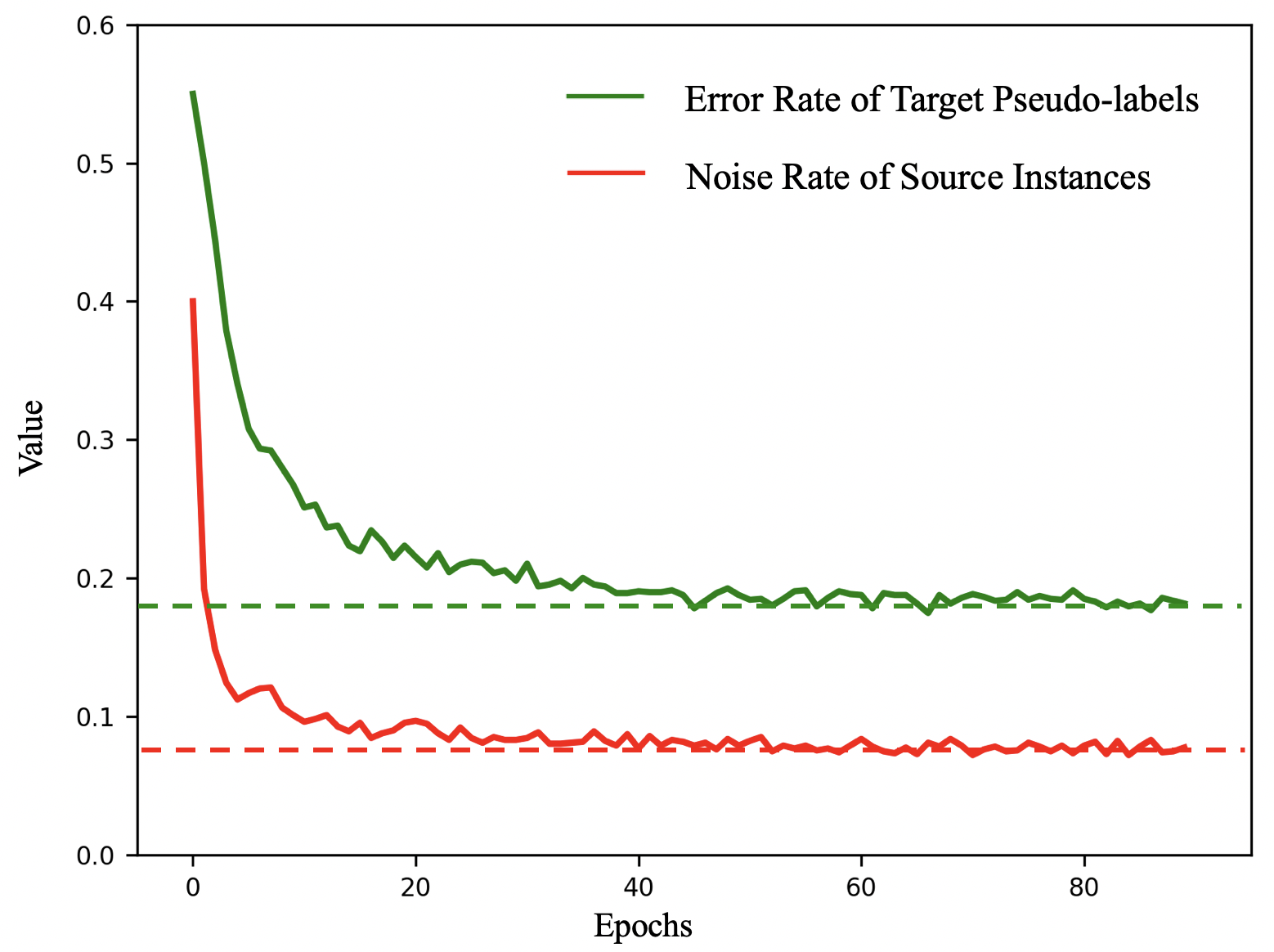}
\caption{Source noise ratio and target pseudo-label error w.r.t training epochs.}
\label{fig5}
\end{figure}

\noindent \textbf{Noise Distribution.} In DualCAN, a semi-supervised deep clustering is adopted to identify feature and label noises. In Fig. 3, we show the separability of feature or label noises from clean instances on Product domain of Office-Home. Specifically, the separability by traditional loss distribution in RDA is shown in Fig. 3 (a), and the separability by clustering in DualCAN is shown in Fig. 3 (b), that is, the separability by distribution of distances from instances to the nearest cluster center. In Fig. 3 (a), there are large overlaps between feature and label noises, indicating that the small-loss criterion cannot clearly separate feature and label noises. While the distance by clustering can better identify feature and label noises, as shown in Fig. 3 (b). Further, we visualize clusters of three random classes in Fig. 3 (c), in which different shapes denote instances in different classes and red triangles denote label-noisy instances that are incorrectly labeled as the other 62 categories. From this figure, it can easily be found that label noises are commonly closer to cluster centers than feature noises, and there are clear boundaries among clusters. It verifies that our proposed NIC can well identify feature and label noises for further noise correction.

\noindent \textbf{Noise Levels.} Fig. 4 reports the performance of individual methods in a wide range of mixed noise levels on \textbf{A}$\rightarrow$\textbf{W} task of Office-31. Specifically, the noise level is from 0.0 to 1.6, where 0.0 means the noise-free UDA scenario, and 1.6 denotes 160$\%$ mix corruption with 80$\%$ label corruption and 80$\%$ feature corruption independently. From Fig. 4, as the noise level increases, the performance decreases for all methods, especially for DANN and ResNet with no consideration of noise processing. The performance of DualCAN is more stable with the increase of noise level, and is better than those of the other methods. It is noted that when the noise level is 1.6, DualCAN performs much better than other methods, including TCL and RDA. The reason can be that DualCAN corrects and recycles noises in learning, thus can make full use of data, especially in high noisy environment. At the same time, DualCAN also achieves the best performance when the noise level is 0, which proves that our proposal is applicable in standard UDA scenarios.

\begin{figure*}[t]
\centering
\subfloat[DANN]{\includegraphics[scale=0.128]{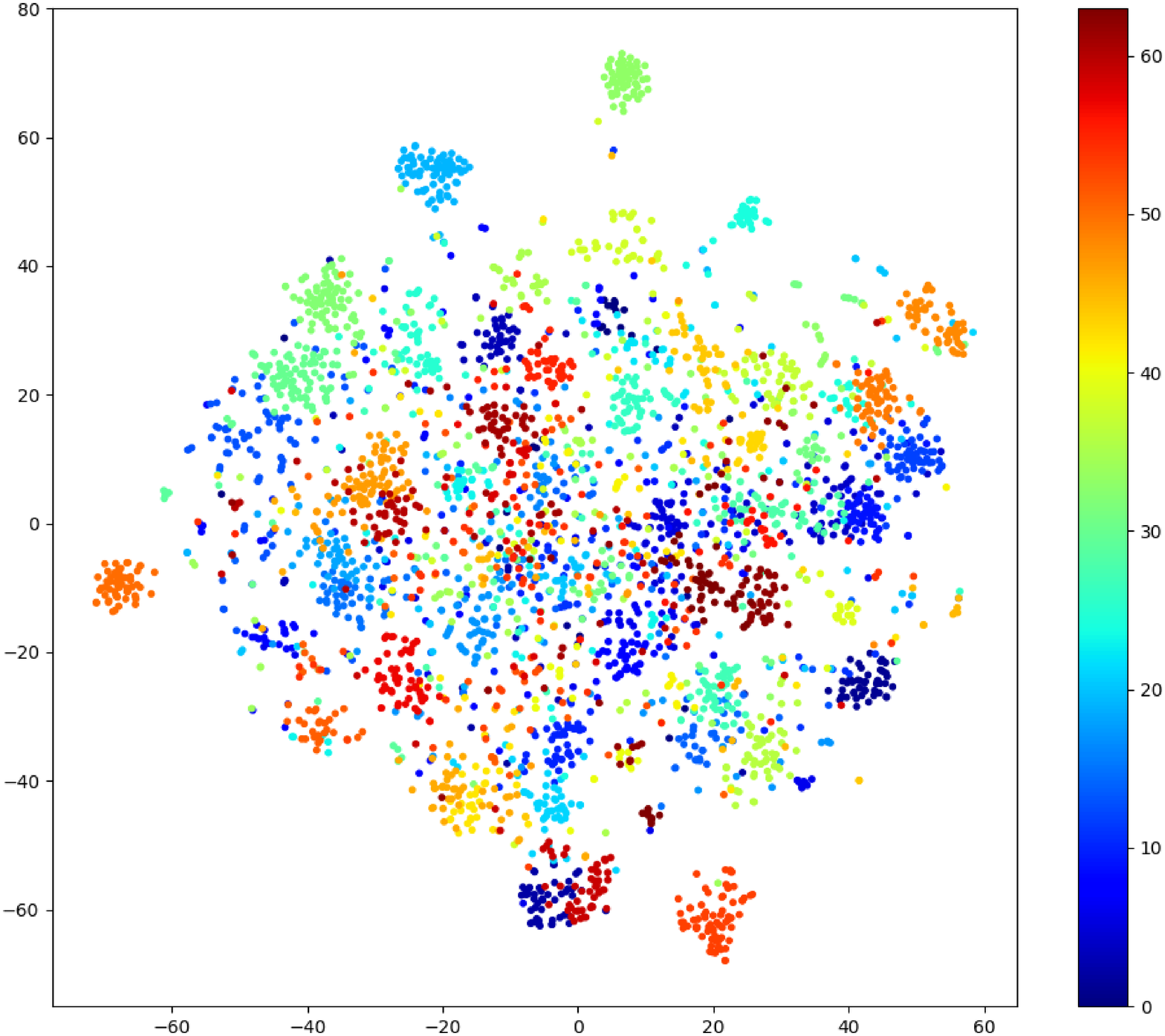}
\label{fig6a}}
\subfloat[TCL]{\includegraphics[scale=0.128]{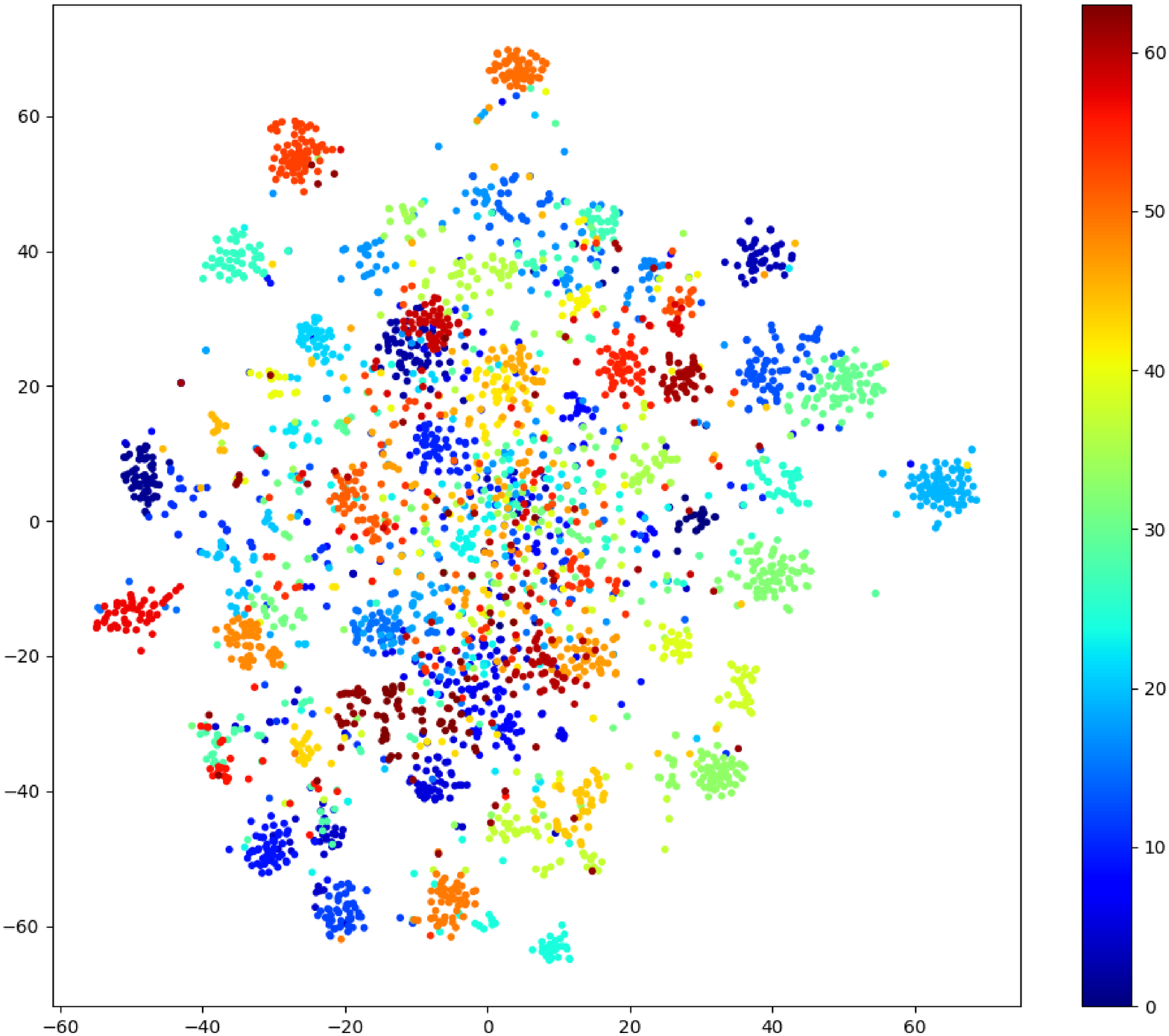}
\label{fig6b}}
\subfloat[RDA]{\includegraphics[scale=0.128]{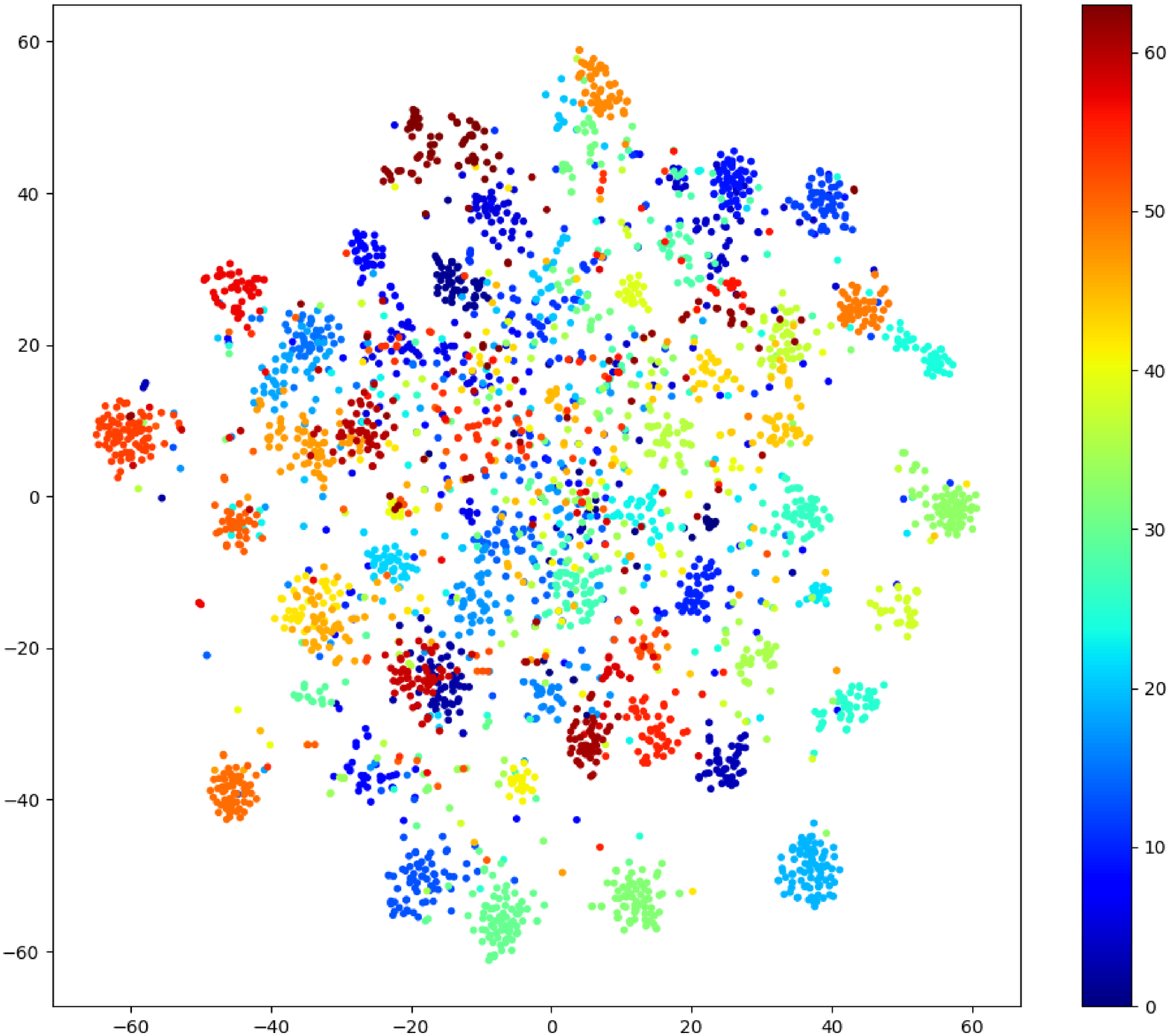}
\label{fig6c}}
\subfloat[DualCAN]{\includegraphics[scale=0.128]{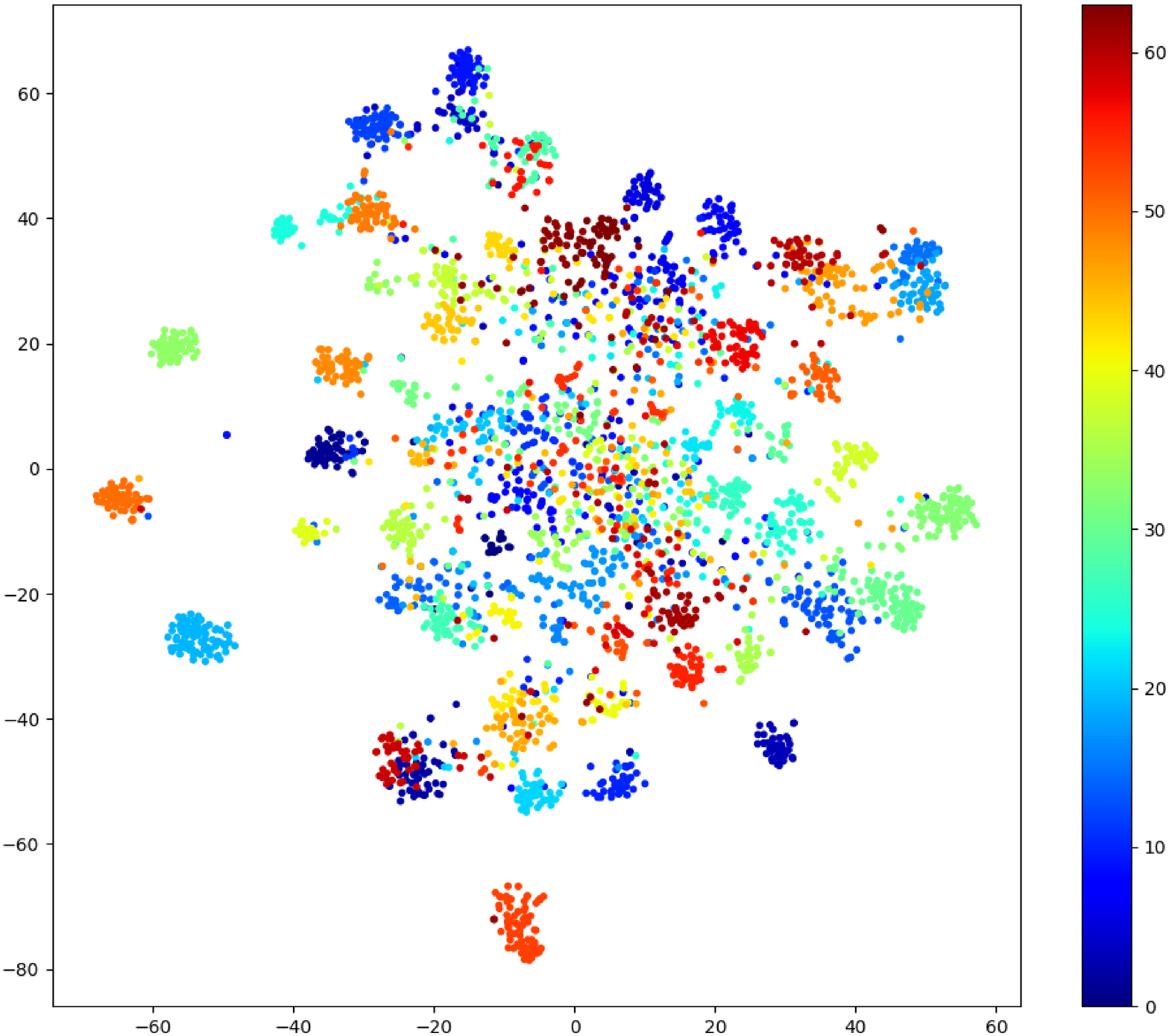}
\label{fig6d}}
\hfil
\subfloat[DANN]{\includegraphics[scale=0.128]{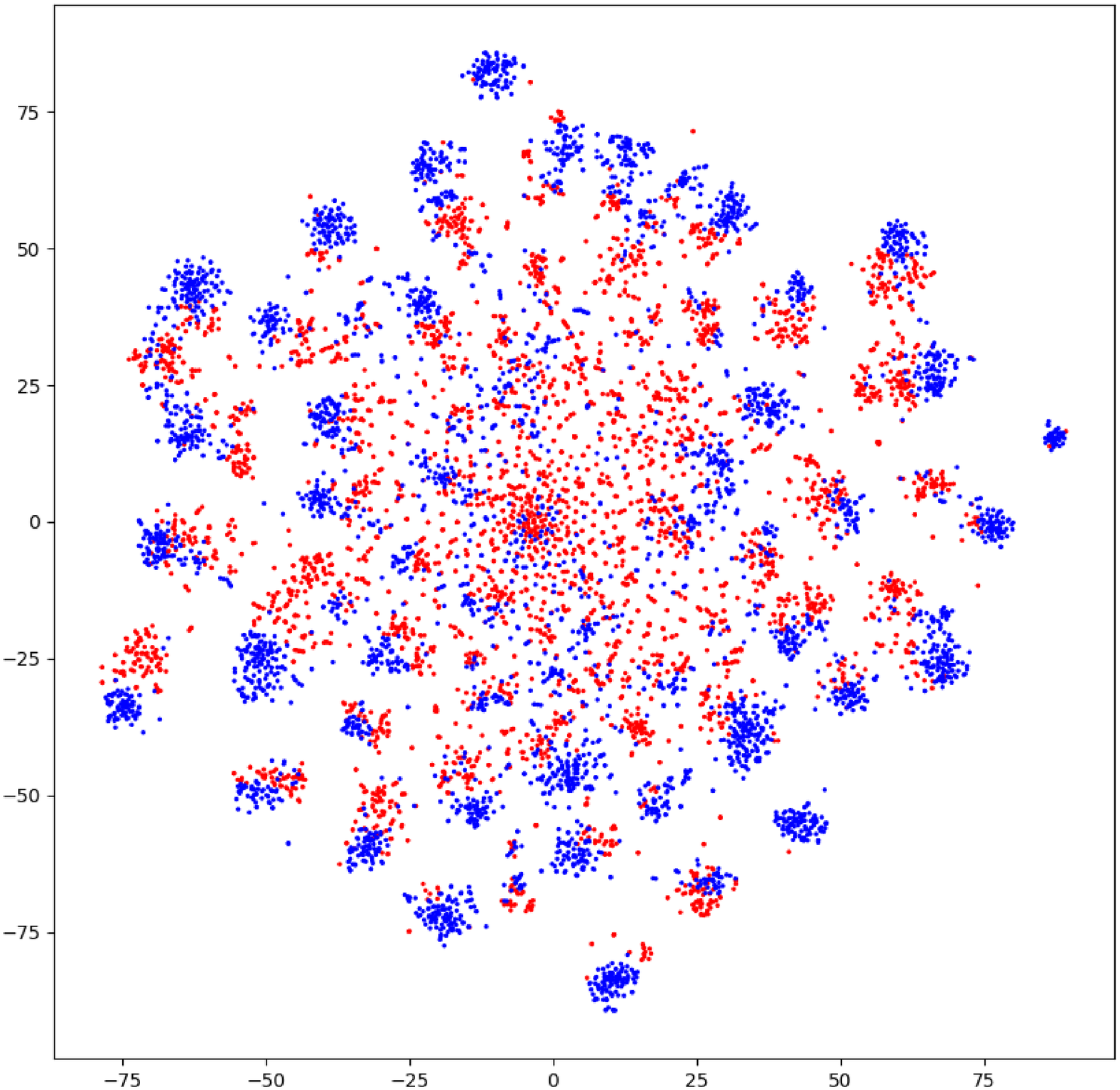}
\label{fig6e}}
\subfloat[TCL]{\includegraphics[scale=0.128]{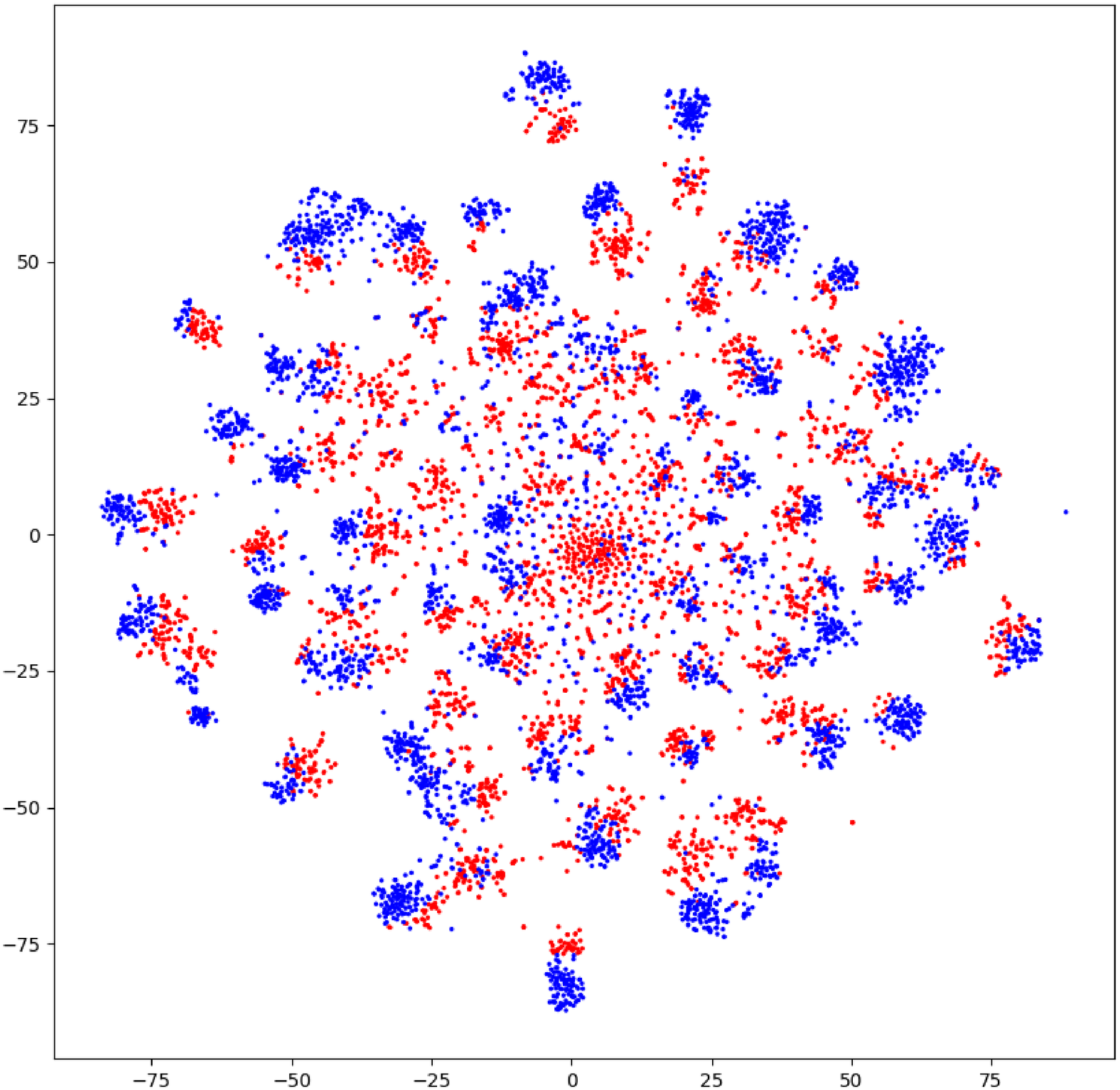}
\label{fig6f}}
\subfloat[RDA]{\includegraphics[scale=0.128]{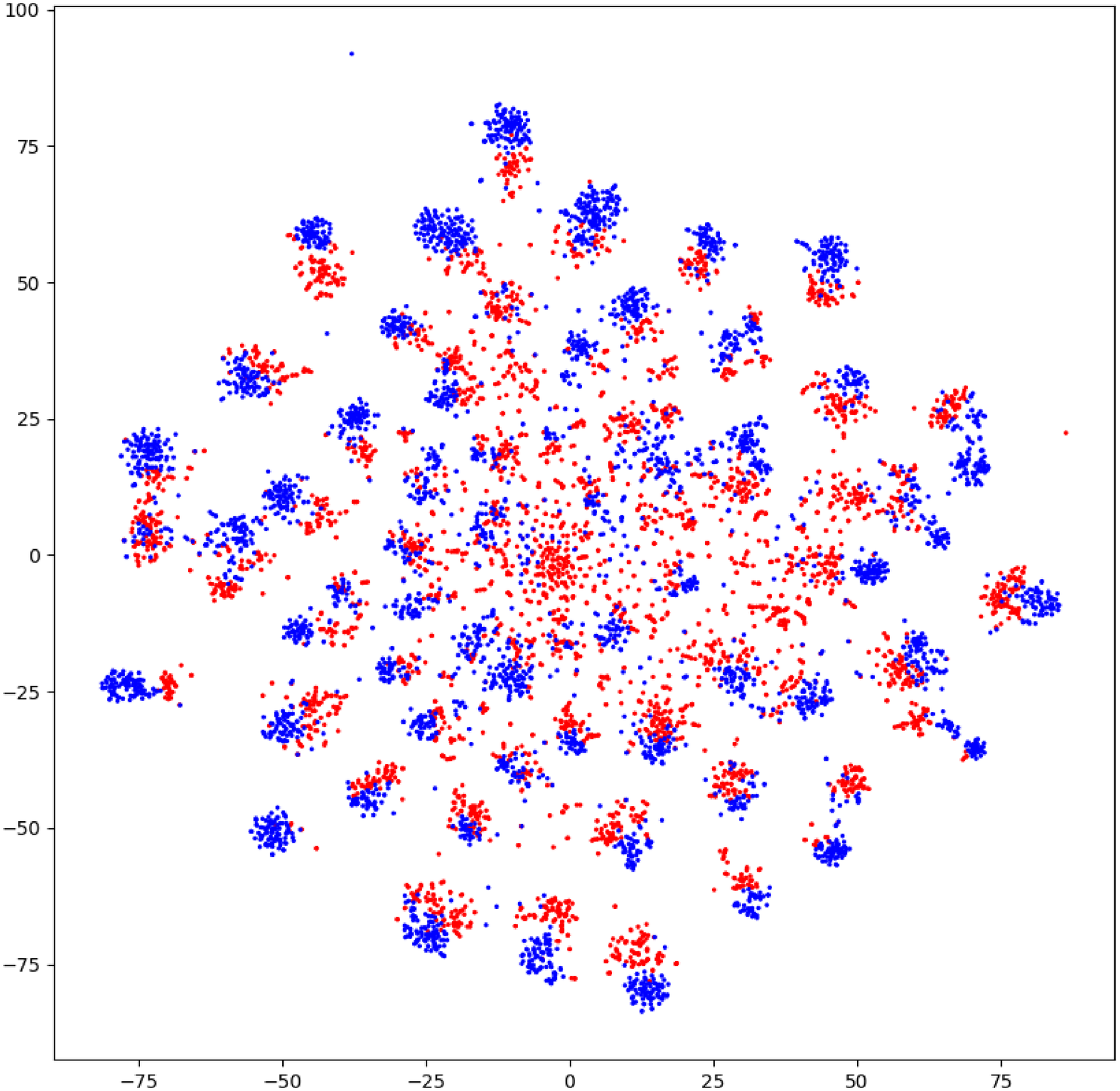}
\label{fig6g}}
\subfloat[DualCAN]{\includegraphics[scale=0.128]{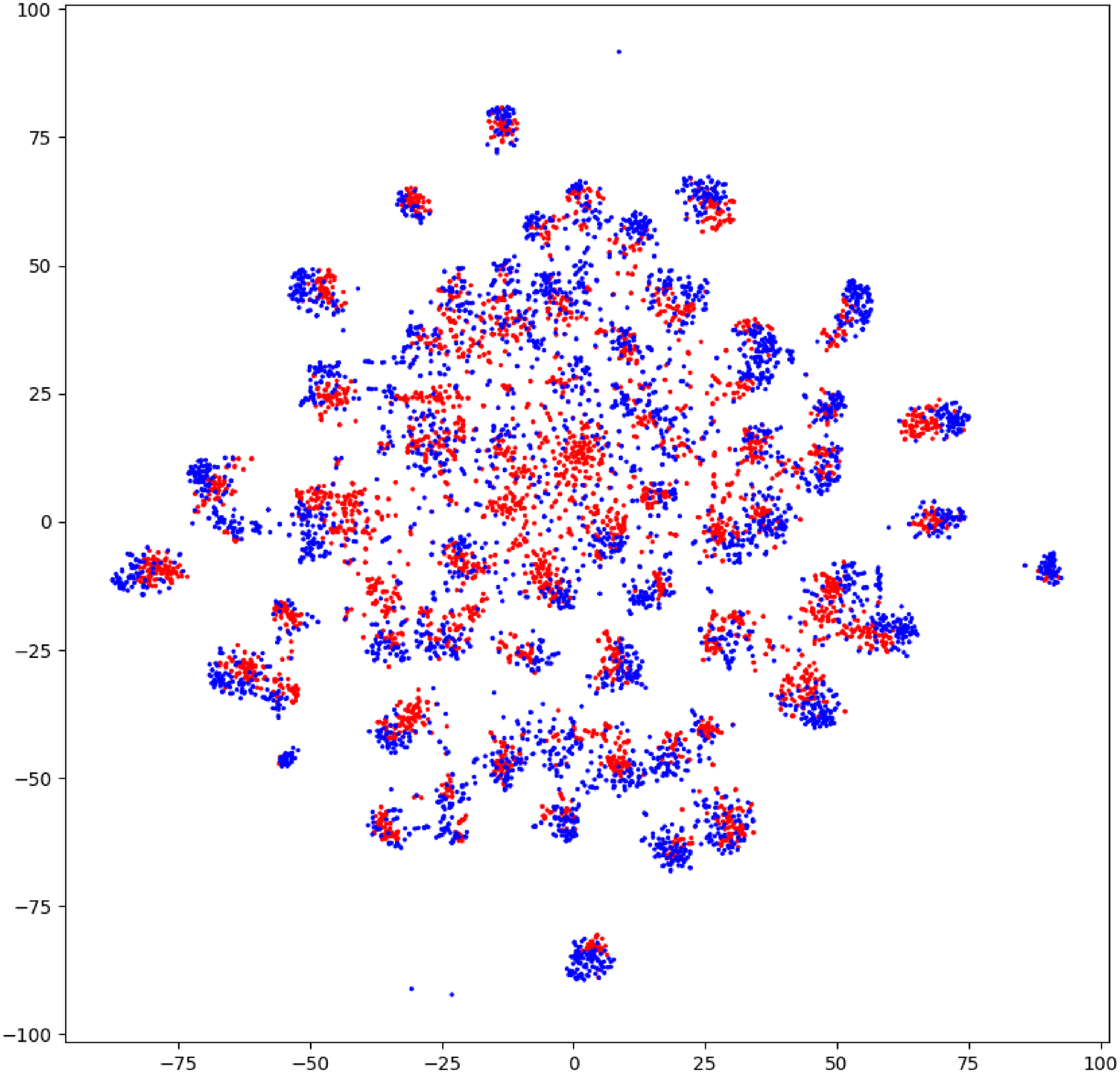}
\label{fig6h}}
\caption{The t-SNE visualization of DANN, TCL, RDA and DualCAN with target class labels (a–d) and domain labels (e–h).}
\label{fig6}
\end{figure*}

\noindent \textbf{Corrected Sample Quality.} In order to show the quality of noise correction, we visualize the source noise ratio and the target pseudo-label error, respectively, on \textbf{Pr}$\rightarrow$\textbf{Rw} task of Office-Home in Fig. 5. As the training proceeds, the proportion of source noise gradually decreases, at the same time, the error of target pseudo-label decays and achieves stable in a few epochs. As a result, with the dual correction and adaptation in DualCAN, the noise in both domains can be corrected mutually, that is, the refined target knowledge can be adopted to correct the source noise, and the corrected source knowledge can further boost the target learning forming a virtuous cycle.

\noindent \textbf{Feature Visualization.} Fig. 6 exhibits the t-SNE embeddings \cite{donahue2014decaf} of the bottleneck representation by DANN, TCL, RDA and DualCAN, respectively, with 40$\%$ mixed corruption on \textbf{Pr}$\rightarrow$\textbf{Rw} of Office-Home. Specifically, Fig. 6 (a)-(c) show the learned target features with different colors indicating different classes, while Fig. 6 (e)-(h) display the learned target features with different colors indicating different domains. From Fig. 6 (a)-(c), the learned features from DANN are mixed up, while DualCAN distinguishes categories better than the other compared methods, including both TCL and RDA, thus it is more robust to noise. At the same time, in Fig. 6 (e)-(h), the source and target domains are better aligned by DualCAN. Moreover, DualCAN obtains clearer boundaries, better within-class compactness, and less outliers. As a result, DualCAN provides a better solution for noisy UDA.

\section{Conclusion}
Prevoius UDA methods commonly focus on the single-directional source-to-target knowledge transfer, while in this paper, a Dual-Correction Adaptation Network (DualCAN) is proposed with dual-directional knowledge transfer and noise correction across domains. In DualCAN, knowledge transfer iterates between source-to-target (ST) task and target-to-source (TS) task. In ST, target pseudo-labels are generated and corrected in terms of source knowledge. While in TS, target knowledge is transferred backward to correct the source noise. Moreover, both feature and label noises are addressed in DualCAN, those noisy instances are corrected and recycled by a Noise Identification and Correction (NIC) module. Empirical results indicate that DualCAN achieves remarkable performance improvement over prior state-of-the-arts.
\section*{Acknowledgments}
This work is supported in part by the NSFC under Grant No.s 62076124 and 62176118.

\ifCLASSOPTIONcaptionsoff
  \newpage
\fi



%
\bibliography{references} 

\begin{thebibliography}{10}
\providecommand{\url}[1]{#1}
\csname url@samestyle\endcsname
\providecommand{\newblock}{\relax}
\providecommand{\bibinfo}[2]{#2}
\providecommand{\BIBentrySTDinterwordspacing}{\spaceskip=0pt\relax}
\providecommand{\BIBentryALTinterwordstretchfactor}{4}
\providecommand{\BIBentryALTinterwordspacing}{\spaceskip=\fontdimen2\font plus
\BIBentryALTinterwordstretchfactor\fontdimen3\font minus
  \fontdimen4\font\relax}
\providecommand{\BIBforeignlanguage}[2]{{%
\expandafter\ifx\csname l@#1\endcsname\relax
\typeout{** WARNING: IEEEtran.bst: No hyphenation pattern has been}%
\typeout{** loaded for the language `#1'. Using the pattern for}%
\typeout{** the default language instead.}%
\else
\language=\csname l@#1\endcsname
\fi
#2}}
\providecommand{\BIBdecl}{\relax}
\BIBdecl

\bibitem{ben2010theory}
S.~Ben-David, J.~Blitzer, K.~Crammer, A.~Kulesza, F.~Pereira, and J.~W.
  Vaughan, ``A theory of learning from different domains,'' \emph{Machine
  learning}, vol.~79, no.~1, pp. 151--175, 2010.

\bibitem{saito2018maximum}
K.~Saito, K.~Watanabe, Y.~Ushiku, and T.~Harada, ``Maximum classifier
  discrepancy for unsupervised domain adaptation,'' in \emph{Proceedings of the
  IEEE conference on computer vision and pattern recognition}, 2018, pp.
  3723--3732.

\bibitem{long2015learning}
M.~Long, Y.~Cao, J.~Wang, and M.~Jordan, ``Learning transferable features with
  deep adaptation networks,'' in \emph{International conference on machine
  learning}.\hskip 1em plus 0.5em minus 0.4em\relax PMLR, 2015, pp. 97--105.

\bibitem{tzeng2017adversarial}
E.~Tzeng, J.~Hoffman, K.~Saenko, and T.~Darrell, ``Adversarial discriminative
  domain adaptation,'' in \emph{Proceedings of the IEEE conference on computer
  vision and pattern recognition}, 2017, pp. 7167--7176.

\bibitem{ganin2016domain}
Y.~Ganin, E.~Ustinova, H.~Ajakan, P.~Germain, H.~Larochelle, F.~Laviolette,
  M.~Marchand, and V.~Lempitsky, ``Domain-adversarial training of neural
  networks,'' \emph{The journal of machine learning research}, vol.~17, no.~1,
  pp. 2096--2030, 2016.

\bibitem{long2018conditional}
M.~Long, Z.~Cao, J.~Wang, and M.~I. Jordan, ``Conditional adversarial domain
  adaptation,'' \emph{Advances in neural information processing systems},
  vol.~31, 2018.

\bibitem{sun2016return}
B.~Sun, J.~Feng, and K.~Saenko, ``Return of frustratingly easy domain
  adaptation,'' in \emph{Proceedings of the AAAI Conference on Artificial
  Intelligence}, vol.~30, no.~1, 2016.

\bibitem{shen2018wasserstein}
J.~Shen, Y.~Qu, W.~Zhang, and Y.~Yu, ``Wasserstein distance guided
  representation learning for domain adaptation,'' in \emph{Thirty-second AAAI
  conference on artificial intelligence}, 2018.

\bibitem{shu2019transferable}
Y.~Shu, Z.~Cao, M.~Long, and J.~Wang, ``Transferable curriculum for
  weakly-supervised domain adaptation,'' in \emph{Proceedings of the AAAI
  Conference on Artificial Intelligence}, vol.~33, no.~01, 2019, pp.
  4951--4958.

\bibitem{han2020towards}
Z.~Han, X.-J. Gui, C.~Cui, and Y.~Yin, ``Towards accurate and robust domain
  adaptation under noisy environments,'' \emph{arXiv preprint
  arXiv:2004.12529}, 2020.

\bibitem{yu2021divergence}
Q.~Yu, A.~Hashimoto, and Y.~Ushiku, ``Divergence optimization for noisy
  universal domain adaptation,'' in \emph{Proceedings of the IEEE/CVF
  Conference on Computer Vision and Pattern Recognition}, 2021, pp. 2515--2524.

\bibitem{zhao2020unsupervised}
F.~Zhao, S.~Liao, G.-S. Xie, J.~Zhao, K.~Zhang, and L.~Shao, ``Unsupervised
  domain adaptation with noise resistible mutual-training for person
  re-identification,'' in \emph{European Conference on Computer Vision}.\hskip
  1em plus 0.5em minus 0.4em\relax Springer, 2020, pp. 526--544.

\bibitem{zellinger2017central}
W.~Zellinger, T.~Grubinger, E.~Lughofer, T.~Natschl{\"a}ger, and
  S.~Saminger-Platz, ``Central moment discrepancy (cmd) for domain-invariant
  representation learning,'' \emph{arXiv preprint arXiv:1702.08811}, 2017.

\bibitem{kang2019contrastive}
G.~Kang, L.~Jiang, Y.~Yang, and A.~G. Hauptmann, ``Contrastive adaptation
  network for unsupervised domain adaptation,'' in \emph{Proceedings of the
  IEEE/CVF Conference on Computer Vision and Pattern Recognition}, 2019, pp.
  4893--4902.

\bibitem{liu2016coupled}
M.-Y. Liu and O.~Tuzel, ``Coupled generative adversarial networks,''
  \emph{Advances in neural information processing systems}, vol.~29, 2016.

\bibitem{hoffman2018cycada}
J.~Hoffman, E.~Tzeng, T.~Park, J.-Y. Zhu, P.~Isola, K.~Saenko, A.~Efros, and
  T.~Darrell, ``Cycada: Cycle-consistent adversarial domain adaptation,'' in
  \emph{International conference on machine learning}.\hskip 1em plus 0.5em
  minus 0.4em\relax PMLR, 2018, pp. 1989--1998.

\bibitem{french2017self}
G.~French, M.~Mackiewicz, and M.~Fisher, ``Self-ensembling for visual domain
  adaptation,'' \emph{arXiv preprint arXiv:1706.05208}, 2017.

\bibitem{lee2013pseudo}
D.-H. Lee \emph{et~al.}, ``Pseudo-label: The simple and efficient
  semi-supervised learning method for deep neural networks,'' in \emph{Workshop
  on challenges in representation learning, ICML}, vol.~3, no.~2, 2013, p. 896.

\bibitem{zou2019confidence}
Y.~Zou, Z.~Yu, X.~Liu, B.~Kumar, and J.~Wang, ``Confidence regularized
  self-training,'' in \emph{Proceedings of the IEEE/CVF International
  Conference on Computer Vision}, 2019, pp. 5982--5991.

\bibitem{kumar2020understanding}
A.~Kumar, T.~Ma, and P.~Liang, ``Understanding self-training for gradual domain
  adaptation,'' in \emph{International Conference on Machine Learning}.\hskip
  1em plus 0.5em minus 0.4em\relax PMLR, 2020, pp. 5468--5479.

\bibitem{cai2021theory}
T.~Cai, R.~Gao, J.~Lee, and Q.~Lei, ``A theory of label propagation for
  subpopulation shift,'' in \emph{International Conference on Machine
  Learning}.\hskip 1em plus 0.5em minus 0.4em\relax PMLR, 2021, pp. 1170--1182.

\bibitem{liu2021cycle}
H.~Liu, J.~Wang, and M.~Long, ``Cycle self-training for domain adaptation,''
  \emph{Advances in Neural Information Processing Systems}, vol.~34, 2021.

\bibitem{miyato2018virtual}
T.~Miyato, S.-i. Maeda, M.~Koyama, and S.~Ishii, ``Virtual adversarial
  training: a regularization method for supervised and semi-supervised
  learning,'' \emph{IEEE transactions on pattern analysis and machine
  intelligence}, vol.~41, no.~8, pp. 1979--1993, 2018.

\bibitem{patrini2017making}
G.~Patrini, A.~Rozza, A.~Krishna~Menon, R.~Nock, and L.~Qu, ``Making deep
  neural networks robust to label noise: A loss correction approach,'' in
  \emph{Proceedings of the IEEE conference on computer vision and pattern
  recognition}, 2017, pp. 1944--1952.

\bibitem{reed2014training}
S.~Reed, H.~Lee, D.~Anguelov, C.~Szegedy, D.~Erhan, and A.~Rabinovich,
  ``Training deep neural networks on noisy labels with bootstrapping,''
  \emph{arXiv preprint arXiv:1412.6596}, 2014.

\bibitem{han2018co}
B.~Han, Q.~Yao, X.~Yu, G.~Niu, M.~Xu, W.~Hu, I.~Tsang, and M.~Sugiyama,
  ``Co-teaching: Robust training of deep neural networks with extremely noisy
  labels,'' \emph{Advances in neural information processing systems}, vol.~31,
  2018.

\bibitem{liu2021co}
J.~Liu, R.~Li, and C.~Sun, ``Co-correcting: noise-tolerant medical image
  classification via mutual label correction,'' \emph{IEEE Transactions on
  Medical Imaging}, vol.~40, no.~12, pp. 3580--3592, 2021.

\bibitem{jiang2018mentornet}
L.~Jiang, Z.~Zhou, T.~Leung, L.-J. Li, and L.~Fei-Fei, ``Mentornet: Learning
  data-driven curriculum for very deep neural networks on corrupted labels,''
  in \emph{International Conference on Machine Learning}.\hskip 1em plus 0.5em
  minus 0.4em\relax PMLR, 2018, pp. 2304--2313.

\bibitem{natarajan2013learning}
N.~Natarajan, I.~S. Dhillon, P.~K. Ravikumar, and A.~Tewari, ``Learning with
  noisy labels,'' \emph{Advances in neural information processing systems},
  vol.~26, 2013.

\bibitem{sukhbaatar2014training}
S.~Sukhbaatar, J.~Bruna, M.~Paluri, L.~Bourdev, and R.~Fergus, ``Training
  convolutional networks with noisy labels,'' \emph{arXiv preprint
  arXiv:1406.2080}, 2014.

\bibitem{chen2021self}
W.~Chen, L.~Lin, S.~Yang, D.~Xie, S.~Pu, Y.~Zhuang, and W.~Ren,
  ``Self-supervised noisy label learning for source-free unsupervised domain
  adaptation,'' \emph{arXiv preprint arXiv:2102.11614}, 2021.

\bibitem{zhu2020deep}
Y.~Zhu, F.~Zhuang, J.~Wang, G.~Ke, J.~Chen, J.~Bian, H.~Xiong, and Q.~He,
  ``Deep subdomain adaptation network for image classification,'' \emph{IEEE
  transactions on neural networks and learning systems}, vol.~32, no.~4, pp.
  1713--1722, 2020.

\bibitem{saenko2010adapting}
K.~Saenko, B.~Kulis, M.~Fritz, and T.~Darrell, ``Adapting visual category
  models to new domains,'' in \emph{European conference on computer
  vision}.\hskip 1em plus 0.5em minus 0.4em\relax Springer, 2010, pp. 213--226.

\bibitem{venkateswara2017deep}
H.~Venkateswara, J.~Eusebio, S.~Chakraborty, and S.~Panchanathan, ``Deep
  hashing network for unsupervised domain adaptation,'' in \emph{Proceedings of
  the IEEE conference on computer vision and pattern recognition}, 2017, pp.
  5018--5027.

\bibitem{bergamo2010exploiting}
A.~Bergamo and L.~Torresani, ``Exploiting weakly-labeled web images to improve
  object classification: a domain adaptation approach,'' \emph{Advances in
  neural information processing systems}, vol.~23, 2010.

\bibitem{he2016deep}
K.~He, X.~Zhang, S.~Ren, and J.~Sun, ``Deep residual learning for image
  recognition,'' in \emph{Proceedings of the IEEE conference on computer vision
  and pattern recognition}, 2016, pp. 770--778.

\bibitem{kumar2010self}
M.~Kumar, B.~Packer, and D.~Koller, ``Self-paced learning for latent variable
  models,'' \emph{Advances in neural information processing systems}, vol.~23,
  2010.

\bibitem{long2016unsupervised}
M.~Long, H.~Zhu, J.~Wang, and M.~I. Jordan, ``Unsupervised domain adaptation
  with residual transfer networks,'' \emph{Advances in neural information
  processing systems}, vol.~29, 2016.

\bibitem{zhang2019bridging}
Y.~Zhang, T.~Liu, M.~Long, and M.~Jordan, ``Bridging theory and algorithm for
  domain adaptation,'' in \emph{International Conference on Machine
  Learning}.\hskip 1em plus 0.5em minus 0.4em\relax PMLR, 2019, pp. 7404--7413.

\bibitem{russakovsky2015imagenet}
O.~Russakovsky, J.~Deng, H.~Su, J.~Krause, S.~Satheesh, S.~Ma, Z.~Huang,
  A.~Karpathy, A.~Khosla, M.~Bernstein \emph{et~al.}, ``Imagenet large scale
  visual recognition challenge,'' \emph{International journal of computer
  vision}, vol. 115, no.~3, pp. 211--252, 2015.

\bibitem{donahue2014decaf}
J.~Donahue, Y.~Jia, O.~Vinyals, J.~Hoffman, N.~Zhang, E.~Tzeng, and T.~Darrell,
  ``Decaf: A deep convolutional activation feature for generic visual
  recognition,'' in \emph{International conference on machine learning}.\hskip
  1em plus 0.5em minus 0.4em\relax PMLR, 2014, pp. 647--655.

\end{thebibliography}
\bibliographystyle{IEEEtran}



%

\begin{IEEEbiography}[{\includegraphics[width=1in,height=1.25in,clip,keepaspectratio]{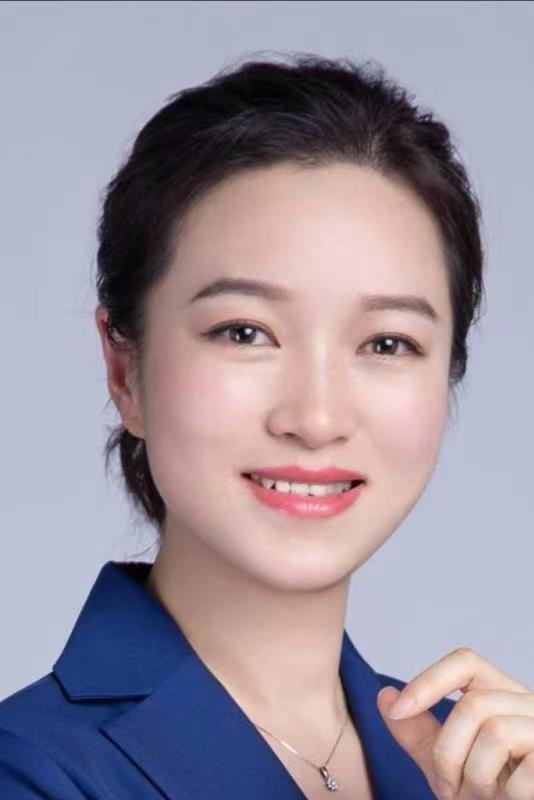}}]{Yunyun Wang}
Yunyun Wang received the Ph.D. degree in Computer Science and Technology from Nanjing University of Aeronautics and Astronautics in 2012. She is currently with the School of Computer Science and Technology in Nanjing University of Posts and Telecommunications. Her current research interests include pattern recognition, machine learning and neural computing.
\end{IEEEbiography}

\begin{IEEEbiography}[{\includegraphics[width=1in,height=1.25in,clip,keepaspectratio]{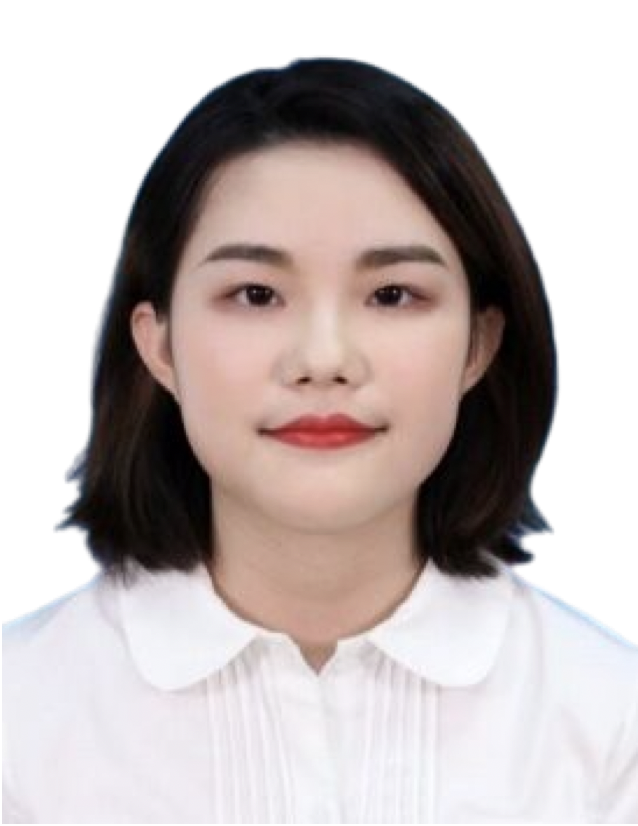}}]{Weiwen Zheng}
Weiwen Zheng is currently working towards the MS degree in computer science at Nanjing University of Posts and Telecommunications, Nanjing, China. Her research interests include deep learning and transfer learning.
\end{IEEEbiography}

\begin{IEEEbiography}[{\includegraphics[width=1in,height=1.25in,clip,keepaspectratio]{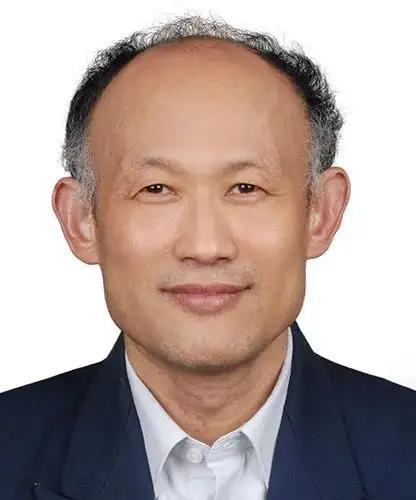}}]{Songcan Chen}
Song-Can Chen received his B.S. degree in mathematics from Hangzhou University (now merged into Zhejiang University), Hangzhou, in 1983. In 1985, he completed his M.S. degree in computer applications at Shanghai Jiao Tong University and then worked at Nanjing University of Aeronautics and Astronautics (NUAA), Nanjing, in January 1986. There he received his Ph.D. degree in communication and information systems in 1997. Since 1998, as a full-time professor, he has been with the College of Computer Science and Technology at NUAA. His research interests include pattern recognition, machine learning, and neural computing. He is also an IAPR Fellow.
\end{IEEEbiography}





\end{document}